\documentclass[11pt]{article}
\usepackage[margin=1in]{geometry}
\usepackage{amsmath,amssymb,amsthm,mathtools}
\usepackage{graphicx,booktabs,array}
\usepackage{enumitem,microtype}
\usepackage[authoryear,round,sort]{natbib}
\usepackage{xcolor}
\usepackage{tikz}
\usetikzlibrary{arrows.meta}
\usepackage[hypertexnames=false,colorlinks=true,linkcolor=blue!55!black,citecolor=blue!55!black,urlcolor=blue!55!black]{hyperref}
\usepackage[capitalize]{cleveref}

\hypersetup{pdftitle={Logarithmic Regret for Alternating Mirror Descent},pdfauthor={Tianlong Nan and Christian Kroer},pdfsubject={Alternating online mirror descent in zero-sum games}}
\setlist{nosep,leftmargin=1.6em}
\newtheorem{theorem}{Theorem}[section]
\newtheorem{lemma}[theorem]{Lemma}
\newtheorem*{formalpreservationrestatement}{Lemma~\ref{lem:formal-modified-energy-preservation}}
\newtheorem*{modifiedenergyrestatement}{Lemma~\ref{lem:modified-energy-tool}}
\newtheorem*{regretdecompositionrestatement}{Lemma~\ref{lem:regret-decomposition}}
\newtheorem*{mainrestatement}{Theorem~\ref{thm:main}}
\newtheorem*{entropyrestatement}{Corollary~\ref{cor:entropy}}
\newtheorem{proposition}[theorem]{Proposition}
\newtheorem{corollary}[theorem]{Corollary}
\theoremstyle{definition}
\newtheorem*{definition}{Definition}
\newtheorem{assumption}[theorem]{Assumption}
\crefname{assumption}{Assumption}{Assumptions}
\Crefname{assumption}{Assumption}{Assumptions}
\theoremstyle{remark}
\newtheorem{remark}[theorem]{Remark}
\newcommand{\R}{\mathbb{R}}

\newcommand{\Gap}{\operatorname{Gap}}
\newcommand{\Reg}{\operatorname{Reg}}
\newcommand{\ri}{\operatorname{ri}}
\newcommand{\dom}{\operatorname{dom}}
\newcommand{\ip}[2]{\langle #1,#2\rangle}
\newcommand{\norm}[1]{\lVert #1\rVert}
\newcommand{\op}{\mathrm{op}}
\newcommand{\dd}{\mathrm{d}}

\title{Logarithmic Regret for Alternating Mirror Descent}
\author{Tianlong Nan$^{\dagger}$ \and Christian Kroer$^{\dagger}$}
\date{}

\begin{document}
\maketitle
\begingroup
\renewcommand{\thefootnote}{\fnsymbol{footnote}}
\footnotetext[2]{Columbia University, Department of Industrial Engineering and Operations Research.\\
Emails: \href{mailto:tianlong.nan@columbia.edu}{\texttt{tianlong.nan@columbia.edu}} and \href{mailto:christian.kroer@columbia.edu}{\texttt{christian.kroer@columbia.edu}}.}
\endgroup
\begin{abstract}
Alternating online mirror descent (AltOMD) is a simple learning rule for zero-sum games, which improves upon simultaneous OMD numerically and theoretically.
Previous work obtained $O(T^{1/3})$ and $O(T^{1/5})$ regret bounds for AltOMD, by leveraging modified-energy-function-based analyses. However, settling whether AltOMD achieves near-constant regret has remained an open question.
In this paper we resolve the question of whether AltOMD achieves near-constant regret in the affirmative.
In particular, we show an $O(\log T)$ regret bound on the sum of the players' external regrets
for any two-player zero-sum bilinear game with compact convex strategy sets, under mild regularity assumptions on the regularizers.
We verify that our assumptions hold for the negative entropy regularizer, 
thereby obtaining logarithmic regret for the alternating multiplicative weights updates algorithm.
The regret bound also yields an $O(\log T/T)$ duality-gap bound for the averaged strategies.
A Lean formalization of the results is available at \url{https://github.com/CoffeeAndConvexity/AltOMD-Lean}.
\end{abstract}

\section{Introduction}
Learning dynamics in games are studied both as models of adaptive strategic behavior and as decentralized algorithms for computing equilibria.
External regret measures each player's performance relative to its best fixed strategy in hindsight.
In two-player zero-sum games, it also connects learning to equilibrium computation: the sum of the players' regrets controls the equilibrium gap of their average strategies.
Sharp regret bounds for simple learning rules therefore help characterize strategic behavior and establish efficient algorithms for solving games.

Alternation is a simple modification of learning dynamics that can substantially improve numerical performance.
Alternating updates exhibit improved stability over simultaneous updates~\citep{gidel2019negative} and are an important component of successful practical game solvers such as CFR+~\citep{tammelin2015solving}.
We study \emph{alternating online mirror descent} (AltOMD), in which the players take turns applying the standard mirror-descent update~\citep{wibisono2022alternating,katona2026symplectic}.
The first player updates against the second player's current strategy, and the second then updates against the first player's new strategy.
Each player uses a regularizer to account for the geometry of its strategy space; negative Shannon entropy on the simplex, for example, yields alternating multiplicative weights update.

A central question is whether AltOMD can achieve near-constant regret in bilinear games with compact convex strategy sets under mild assumptions on the regularizers.
In particular, an $O(\log T)$ bound on the sum of the players' regrets over $T$ rounds would yield an $O(\log T/T)$ equilibrium gap for their average strategies.

\citet*{wibisono2022alternating} connected AltOMD to Hamiltonian dynamics in the dual space.
By controlling the first correction to the energy under third-order smoothness, they obtained $O(T^{1/3})$ total regret and an $O(T^{-2/3})$ equilibrium gap for the average strategies.
\citet*{katona2026symplectic} extended this approach using a modified Hamiltonian constructed from the Baker--Campbell--Hausdorff (BCH) series, improving these rates to $O(T^{1/5})$ and $O(T^{-4/5})$, respectively.
\citet*{katona2026symplectic} further conjectured estimates at arbitrary order that would imply $O(T^\varepsilon)$ regret for every fixed $\varepsilon>0$.
They also established bounded regret under additional convergence and tail conditions on the modified Hamiltonian, or boundedness of the dual trajectories.
A logarithmic regret guarantee under mild regularizer assumptions has nevertheless remained open, including for alternating multiplicative weights on the simplex.

The key obstacle is controlling the error when the modified Hamiltonian is truncated at an order that grows with the horizon $T$.
Estimates at each fixed order do not suffice: their hidden constants may grow too quickly with the order.
Moreover, the infinite modified-Hamiltonian series need not converge, and compact strategy sets do not ensure bounded dual trajectories.
Extending this approach to logarithmic regret therefore requires explicit control of higher-order corrections throughout the dual space, as discussed by~\citet[Section~9.1]{katona2026symplectic}.

\paragraph{Our result.}
We prove an $O(\log T)$ bound on the sum of the players' external regrets for AltOMD in two-player zero-sum bilinear games with compact convex strategy sets~(\cref{thm:main}).
The result applies to Legendre regularizers whose conjugates satisfy a uniform geometric bound on their Taylor coefficients, assuming a finite initial Bregman radius.
For a prescribed horizon $T$, the algorithm uses a stepsize $\eta_T=\Theta(1/\log T)$ that remains constant throughout the run.
The average strategies then have duality gap $O(\log T/T)$.

These assumptions hold for negative Shannon entropy, yielding the same guarantees for alternating multiplicative weights on the simplex~(\cref{cor:entropy}).
Our result improves the conjectured $O(T^\varepsilon)$ target to logarithmic regret for this class of regularizers and applies even when equilibria lie on the boundary and dual trajectories are unbounded.

\subsection{Related work}

\paragraph{Alternation in game dynamics.}
Alternation can substantially change the stability and convergence properties of learning dynamics.
Early results concern unconstrained bilinear games, where simultaneous GDA diverges.
\citet{gidel2019negative} showed that alternating gradient descent-ascent (AltGDA) instead keeps the iterates bounded, though it does not converge, 
and that adding negative momentum yields linear convergence.
\citet{bailey2020finite} show that a player using alternating gradient descent has bounded regret against every fixed comparator, whatever the opponent does, 
under a regret notion that evaluates each opponent strategy against the player's iterates both before and after its update.

For constrained games, 
\citet{wibisono2022alternating} connected AltOMD to Hamiltonian dynamics in the dual space and 
obtained $O(T^{1/3})$ total regret through a modified-energy argument.
They also highlighted that the discretized AltOMD algorithm inherits some dynamic properties from continuous-time dynamics~\citep{mertikopoulos2018cycles}.
\citet{katona2026symplectic} developed higher-order modified-Hamiltonian expansions and 
improved this bound to $O(T^{1/5})$ under stronger smoothness assumptions.
For projected AltGDA, 
\citet{nan20261} established $O(1/T)$ ergodic convergence under an interior-equilibrium assumption and 
a local $O(1/T)$ ergodic guarantee in general matrix games.
\citet{nan2026global} subsequently proved global $O(1/T)$ ergodic convergence for all finite two-player zero-sum matrix games.
The effects of alternation have also been explored in self-play settings~\citep{tammelin2015solving,burch2019revisiting,cai2025last,li2026why,sullins2026ir}.

A complementary line studies alternating regret against adversarial losses, 
developing algorithms tailored to this criterion for online linear and convex optimization~\citep{cevher2023alternation,hait2025alternating}.
Recent independent work by~\citet{tao2026optimal} and~\citet{zhang2026minimax} obtains
optimal alternating-regret bounds using modified exponential-weights algorithms.
In particular, their modified algorithms guarantee an $O(\log d/T)$ averaged duality gap in finite two-player zero-sum games.
Our work, in contrast, uses the standard unmodified AltOMD dynamics, showing that no modification is necessary to achieve such a rate.

\paragraph{Learning dynamics in matrix games.}
No-regret learning provides a general route to equilibrium computation.
Standard no-regret learning dynamics can yield $O(1/\sqrt{T})$ average-iterate convergence rate in two-player zero-sum games,
while faster convergence rates are available through predictive or corrective updates.
Mirror-Prox achieves an $O(1/T)$ averaged duality gap for smooth convex-concave saddle-point problems \citep{nemirovski2004prox}.
\citet{rakhlin2013optimization} develop optimistic mirror descent using predictable sequences and obtain $O(\log T/T)$ averaged-iterate convergence for strongly uncoupled learning in matrix games.
\citet{syrgkanis2015fast} further shows
a constant bound on the sum of regrets for suitable optimistic updates with constant stepsizes,
implying an $O(1/T)$ average-iterate convergence rate in zero-sum games.

Recent work also sharpens individual-regret guarantees in finite general-sum games.
\citet{ha2026logarithmic} proves $O(\sqrt{n}\log d_i\log T)$ individual external regret for Optimistic Hedge~\citep{rakhlin2013optimization},
where $n$ is the number of players and $d_i$ is player $i$'s number of actions, 
with a common stepsize of order $1/(\sqrt{n}\log T)$ held constant during play.
Other recent algorithms obtain horizon-independent individual external regret through 
exponential-moving-average predictors and lifted regularization \citep{liu2026constant}, 
discounted higher-order optimism \citep{abbadi2026constant}, 
and one-step multiplicatively optimistic regret matching \citep{soleymani2026multiplicative}.
For the stronger swap-regret criterion, \citet{tsuchiya2026sublogarithmic} obtains $O(\sqrt{\log T})$ regret for fixed game dimensions by combining the Blum-Mansour reduction with hybrid-regularized optimistic FTRL.

\section{Preliminaries}
\label{sec:setup}

We study the bilinear two-player zero-sum game defined by the payoff matrix $A\in\R^{m\times n}$ and a pair of nonempty compact and convex strategy sets $\mathcal{X} \subset \R^n$ and $\mathcal{Y} \subset \R^m$.
In this game, 
the first player chooses her strategy $x$ to minimize the bilinear payoff,
while the second player chooses her strategy $y$ to maximize it.
The game can then be encoded as the following minimax problem: 
\begin{equation}
    \min_{x\in \mathcal{X}}\max_{y\in \mathcal{Y}} y^\top A x.
    \tag{Bilinear game}
    \label{eq:game}
\end{equation}

\paragraph{Legendre regularizer and dual space.}
Let $\psi_{\mathcal{X}}$ and $\psi_{\mathcal{Y}}$ be proper closed convex functions 
whose domains have closures $\mathcal{X}$ and $\mathcal{Y}$, respectively.
We takes their values to be $+\infty$ outside their domains. 
In this paper, we call them \emph{regularizers} or \emph{mirror maps}.
We consider the Legendre regularizers defined in the following definition.
We use $\ri(\cdot)$ to denote the relative interior of a set.
\begin{definition}[Legendre functions]
A proper closed convex function $\psi:\R^d\to\R\cup\{+\infty\}$ is \emph{Legendre}
if it is strictly convex and continuously differentiable on $\ri(\dom\psi)$,
and $\norm{\nabla\psi(x_k)}_2\to\infty$ whenever a sequence
$x_k\in\ri(\dom\psi)$ converges to a relative boundary point of $\dom\psi$.
Here, gradients and boundaries are taken relative to the affine hull of the domain of $\psi$.
\end{definition}

Then, we define the convex conjugates of the regularizer functions by
\begin{equation}
    \begin{aligned}
        F(u) &:= \psi_{\mathcal{X}}^*(u) = \sup\nolimits_{x\in \mathcal{X}} \left\{\ip{u}{x}-\psi_{\mathcal{X}}(x)\right\}, \\
        G(w) &:= \psi_{\mathcal{Y}}^*(w) = \sup\nolimits_{y\in \mathcal{Y}} \left\{\ip{w}{y}-\psi_{\mathcal{Y}}(y)\right\}.
    \end{aligned}
    \label{eq:conjugates}
\end{equation}

Throughout this paper, we call the joint feasible set $\mathcal{X}\times\mathcal{Y}$ the (product) \emph{primal space}.
To facilitate the analysis, we consider an associated (product) \emph{dual space} or mirror space $\R^n\times\R^m$ with coordinates $z := (u,w)$.
Given Legendre regularizers $\psi_{\mathcal{X}}$ and $\psi_{\mathcal{Y}}$, 
Each pair of dual variables $u$ and $w$ can be mapped to a corresponding pair of primal strategies $x$ and $y$ by
\begin{equation}
    x = \nabla F(u) \in \ri(\mathcal{X}), \qquad y = \nabla G(w) \in \ri(\mathcal{Y}).
\end{equation}

\paragraph{Alternating mirror descent (AltOMD).}
Consider a Legendre regularizer $\psi$ on $\R^d$. 
For $x' \in\dom\psi$ and $x\in\ri(\dom\psi)$, the \emph{Bregman divergence}
generated by $\psi$ is defined as
\begin{equation*}
    D_\psi(x',x):=\psi(x')-\psi(x)-\ip{\nabla\psi(x)}{x'-x} \geq 0, 
\end{equation*}
where the nonnegativity follows from the convexity of $\psi$. 
For $x'\notin\dom\psi$, we set $D_\psi(x',x):=+\infty$.
The most common example of a Bregman divergence is the KL divergence, which arises from the negative entropy regularizer.

The alternating online mirror descent (AltOMD) updates in the primal space are given by
\begin{equation}
    \begin{aligned}
        x_{k+1} &= \arg\min\nolimits_{x\in\mathcal{X}} \left\{\eta\ip{A^\top y_k}{x}+D_{\psi_{\mathcal{X}}}(x,x_k)\right\}, \\
        y_{k+1} &= \arg\min\nolimits_{y\in\mathcal{Y}} \Big\{ -\eta\ip{A x_{k+1}}{y}+D_{\psi_{\mathcal{Y}}}(y,y_k) \Big\},
    \end{aligned}
    \tag{AltOMD-Primal}
    \label{eq:altomd-primal}
\end{equation}
where $\eta>0$ is the step size.
Equivalently, by the first-order optimality conditions, the updates can be expressed in the dual space as follows:
\begin{equation}
    \begin{aligned}
        u_{k+1}&=u_k-\eta A^\top y_k,& x_{k+1}&=\nabla F(u_{k+1}),\\
        w_{k+1}&=w_k+\eta A x_{k+1},& y_{k+1}&=\nabla G(w_{k+1}).
    \end{aligned}
    \tag{AltOMD-Dual}
    \label{eq:altomd-dual}
\end{equation}
Alternation means that the players update their strategies one after another, rather than simultaneously: 
the first player updates $x_{k+1}$ based on the previous strategy of the second player $y_k$, 
and then the second player updates $y_{k+1}$ using the newly updated $x_{k+1}$.

Since the AltOMD updates in the dual space are unconstrained and additive, it is convenient to analyze the algorithm in the dual space. 

\paragraph{Continuous-time Hamiltonian dynamics.}
Similar to the previous work~\citep{wibisono2022alternating,katona2026symplectic},
we leverage Hamiltonian dynamics to analyze AltOMD.
Here, we present a brief overview of continuous-time Hamiltonian dynamics.

Hamiltonian dynamics describe continuous-time dynamics whose evolution is driven by a scalar energy function.
In the AltOMD case, the continuous-time Hamiltonian dynamics corresponding to~\eqref{eq:altomd-dual} is
\begin{equation}
    \dot u(t)=-A^\top\nabla G(w(t)),
    \qquad
    \dot w(t)=A\nabla F(u(t)).
    \label{eq:continuous-dynamics}
\end{equation}
Set $z:=(u,w)$. Then~\cref{eq:continuous-dynamics} takes the compact form $\dot z=J_A\nabla H_0(z)$, 
where 
\begin{equation}
    H_0(u,w) := F(u) + G(w) \ 
    \mbox{ and } \ 
    J_A:=\begin{pmatrix}0&-A^\top\\ A&0\end{pmatrix}.
    \label{eq:poisson-matrix}
\end{equation}
Since $J_A^\top=-J_A$, the energy is conserved along the continuous trajectory:
\begin{equation*}
    \frac{\dd}{\dd t}H_0(z(t)) 
    = \nabla H_0(z(t))^\top \dot{z}(t)  
    = \nabla H_0(z(t))^\top J_A \nabla H_0(z(t)) = 0.
\end{equation*}

AltOMD approximates the continuous-time Hamiltonian dynamics by sequentially updating the two variables.

\paragraph{Regret and duality gap.}
We define the regret up to time $T$ as the sum of the minimizing and maximizing players' external regrets:
\begin{align}
    \Reg_T
    &:= \max_{\tilde{x}\in\mathcal{X}}\sum\nolimits_{k=0}^{T-1}y_k^\top A(x_{k+1}-\tilde{x})
    +\max_{\tilde{y}\in\mathcal{Y}}\sum\nolimits_{k=0}^{T-1}(\tilde{y}-y_k)^\top Ax_{k+1}
    \notag \\ 
    &= \max_{(\tilde{x},\tilde{y})\in\mathcal{X}\times\mathcal{Y}}
    \sum\nolimits_{k=0}^{T-1}\bigl(\tilde{y}^\top Ax_{k+1}-y_k^\top A\tilde{x}\bigr).
    \label{eq:regret}
\end{align}

Let $\bar{x}_T := \frac{1}{T} \sum_{k=0}^{T-1} x_{k+1}$ and $\bar{y}_T := \frac{1}{T} \sum_{k=0}^{T-1} y_k$.
For any strategy pair $(x, y) \in \mathcal{X} \times \mathcal{Y}$, 
the duality gap is 
\begin{equation}
    \Gap(x,y) := \max_{\tilde{y}\in\mathcal{Y}}\tilde{y}^\top Ax
    -\min_{\tilde{x}\in\mathcal{X}}y^\top A\tilde{x}.
    \label{eq:gap}
\end{equation}
The duality gap is nonnegative and vanishes exactly when $(x,y)$ is a saddle point of~\eqref{eq:game}.
By bilinearity, we have 
\begin{equation}
    \Reg_T 
    = T \left(\max\nolimits_{\tilde{y} \in \mathcal{Y}} \tilde{y}^\top A \bar{x}_T
        - \min\nolimits_{\tilde{x} \in \mathcal{X}} \bar{y}_T^\top A \tilde{x} \right) 
    = T\Gap(\bar x_T,\bar y_T).
    \label{eq:regret-gap}
\end{equation}
Hence, a sublinear regret implies that the duality gap of the averaged strategy converges to zero.

\section{Logarithmic Regret Bounds}
    \label{sec:main}
In this section, we establish logarithmic regret bounds for the AltOMD algorithm.

\subsection{Notation and Assumptions}
We first introduce some additional notation and assumptions.
For a smooth scalar function $f: \R^d \to \R$ and an integer $s \geq 1$,
let $D^s f(z)$ denote its $s$-th derivative at $z$, 
which is an order-$s$ tensor.
For an order-$s$ tensor $T$, we use $T[v_1,\ldots,v_s]$ to denote its contraction with the vectors $v_1,\ldots,v_s$.
For example, for $s=1$, $D^1 f(z)[v_1] = \nabla f(z)^\top v_1$, and for $s=2$, $D^2 f(z)[v_1,v_2] = v_1^\top \nabla^2 f(z) v_2$.
For any integer $s \geq 1$, we define the operator norm induced by Euclidean norms as 
\begin{equation}
    \norm{D^sf(z)}_{\op} := \sup_{\norm{v_1}_2,\ldots,\norm{v_s}_2\leq 1}
    \bigl|D^sf(z)[v_1,\ldots,v_s]\bigr|. 
\end{equation}
For $s=1$ and $s=2$, 
the operator norms reduce to the Euclidean norm and spectral norm, respectively.

The first assumption imposes a derivative growth condition on the conjugate functions $F = \psi_{\mathcal{X}}^*$ and $G = \psi_{\mathcal{Y}}^*$.
This condition ensures that the growth of derivatives is controlled at \emph{every} order,
which is crucial for the subsequent analysis.
\begin{assumption}[Derivative growth condition]
    \label{ass:regularity}
    There is a constant $K \geq 1$ such that, for every integer $s\geq 1$,
    \begin{equation}
        \frac{1}{s!} \left( \sup_u\norm{D^sF(u)}_{\op}+\sup_w\norm{D^sG(w)}_{\op} \right) \leq K^s.
        \label{eq:factorial}
    \end{equation}
\end{assumption}
We comment that this assumption is mild in the sense that it holds for common regularizers 
such as the negative entropy regularizer, which is shown in~\cref{sec:examples}.

The second assumption is standard and ensures that the initial Bregman divergence is finite.
For example, when the feasible sets are simplices and the regularizers are negative entropy, 
both players can choose the initial strategies to be the uniform distributions over the strategy sets.
\begin{assumption}[Finite initial comparator radius]\label{ass:radius}
The initial strategy $(x_0, y_0)$ satisfies that
\begin{equation}\label{eq:radius}
    \Lambda_0 := \sup\nolimits_{\tilde{x}\in\mathcal{X}}D_{\psi_{\mathcal{X}}}(\tilde{x},x_0) + \sup\nolimits_{\tilde{y}\in\mathcal{Y}}D_{\psi_{\mathcal{Y}}}(\tilde{y},y_0)<\infty.
\end{equation}
\end{assumption}

\subsection{Analysis}

We begin with the following regret decomposition.
All the proofs in this section are deferred to~\cref{app:proof}.
\begin{lemma}[Regret decomposition]\label{lem:regret-decomposition}
Fix a time horizon $T \geq 1$ and a step size $\eta > 0$.
For any comparators $(\tilde{x},\tilde{y}) \in \dom\psi_{\mathcal{X}}\times\dom\psi_{\mathcal{Y}}$,
define the Fenchel coupling: 
\begin{equation*}
    \Gamma_{\tilde{x},\tilde{y}}(u,w)
    :=
    F(u) - \left( \ip{u}{\tilde{x}} - \psi_{\mathcal{X}}(\tilde{x}) \right) 
    + G(w) - \left( \ip{w}{\tilde{y}} -  \psi_{\mathcal{Y}}(\tilde{y}) \right) 
    \geq 0,
\end{equation*}
where $(u, w)$ are the dual variables in~\eqref{eq:altomd-dual}.
Then, 
we have $\Gamma_{\tilde{x},\tilde{y}}(u_k, w_k) = D_{\psi_{\mathcal{X}}}(\tilde{x},x_k) + D_{\psi_{\mathcal{Y}}}(\tilde{y},y_k)$ for all $k \geq 0$.
Moreover, all the iterates generated by the AltOMD updates satisfy 
\begin{equation}\label{eq:comparator-identity}
    \eta \bigl(
        \tilde{y}^\top A x_{k+1} - y_k^\top A \tilde{x}
    \bigr)
    =
    H_0(z_{k+1}) - H_0(z_k)
    + \Gamma_{\tilde{x},\tilde{y}}(z_k) - \Gamma_{\tilde{x},\tilde{y}}(z_{k+1}),
\end{equation}
where $H_0(z) = F(u) + G(w)$ is defined in~\cref{eq:poisson-matrix}.
\end{lemma}

We sum the right-hand side of~\cref{eq:comparator-identity} over $k=0,\dots,T-1$.
The intermediate terms cancel, so only the boundary terms remain:
$H_0(z_T) - H_0(z_0)$ and $\Gamma_{\tilde{x},\tilde{y}}(z_0) - \Gamma_{\tilde{x},\tilde{y}}(z_T)$. 
The second term $\Gamma_{\tilde{x},\tilde{y}}(z_0) - \Gamma_{\tilde{x},\tilde{y}}(z_T)$ can be bounded by the nonnegativity of Bregman divergences and the initial strategy condition~\cref{ass:radius}.

The difficulty lies in controlling the first term $H_0(z_T) - H_0(z_0)$.
As mentioned earlier, 
this $H_0$ corresponds to the original energy in the Hamiltonian dynamics, 
and is preserved under the continuous-time counterpart of the AltOMD.
For this reason, 
the corresponding continuous-time dynamics achieves a constant regret 
because $H_0(z_T) = H_0(z_0)$ for all $T \geq 0$.
However, 
discretized updates of the continuous-time dynamics generally do not preserve this energy, 
and there is no guarantee that $H_0(z_T)$ will be bounded above.

To handle this issue, 
following \citet{katona2026symplectic},
we leverage a modified Hamiltonian energy function~\citep{field2003note}. %
This modified energy need not converge, and because of this we treat it as a formal power series~\citep{niven1969formal}: its coefficients are manipulated algebraically without assuming convergence.

We write the modified energy in the following formal power series form: 
\begin{equation}
    \widehat{H}_\eta(z) := H_0(z) + \eta H_1(z) + \eta^2 H_2(z) + \cdots,
    \label{eq:formal-modified-energy-sketch}
\end{equation}
where each correction term $H_r$, $r\geq1$ is
a linear combination of nested/iterated Poisson brackets $\{F,G\}$ of $F$ and $G$, which is defined as 
\begin{equation*}
    \label{eq:poisson}
    \{F,G\} := \nabla F^\top J_A\nabla G.
    \tag{Poisson bracket}
\end{equation*}
For example, the first and second correction terms are 
\begin{align*}
    H_1(z) &= \frac{1}{2} \{F,G\}(u,w)
    = -\frac{1}{2}\nabla G(w)^\top A\nabla F(u), \\ 
    H_2(z) &= \frac1{12}\{G,\{G,F\}\}(u,w) + \frac1{12}\{F,\{F,G\}\}(u,w).
\end{align*}
We defer the complete construction of the correction terms to~\cref{app:bch}.

We further define $\phi_G^\eta(u,w) := \bigl(u-\eta A^\top\nabla G(w), w\bigr)$, 
$\phi_F^\eta(u,w) := \bigl(u, w+\eta A\nabla F(u)\bigr)$ and 
$\mathcal M_\eta = \phi_F^\eta\circ\phi_G^\eta$,
where $\circ$ denotes function composition.
Hence, by~\eqref{eq:altomd-dual}, the AltOMD updates correspond to applying the map $\mathcal M_\eta$ to the current iterate, 
i.e., $z_{k+1}=\mathcal M_\eta(z_k)$.
With this notation, we present 
the following classical formal preservation identity, 
which is proved in, e.g.,~\citet[Chapters~III and~IX]{hairer2006geometric}.
For completeness, we provide a self-contained proof in~\cref{app:bch}.
Here, the formal preservation refers to a coefficient-by-coefficient equality in the formal power series.
\begin{lemma}[Formal preservation of $\widehat{H}_\eta$]
    \label{lem:formal-modified-energy-preservation}
    Let $F$ and $G$ be smooth, let $H_r$ be the BCH correction terms constructed in~\cref{app:bch}, and let $\widehat H_\eta$ be the formal modified energy defined by~\cref{eq:formal-modified-energy-sketch}.
    Then, $\widehat H_\eta$ satisfies
    \begin{equation}
        \widehat H_\eta\bigl(\mathcal M_\eta(z)\bigr) = \widehat H_\eta(z)
        \qquad\text{for every } z \in \R^n \times \R^m,
    \label{eq:formal-modified-energy-preservation}
\end{equation}
    as an identity of formal power series in $\eta$.
\end{lemma}

For any integer $N \geq 0$, we consider an \emph{arbitrary-order truncation} for the modified energy.
Given any integer $N\geq0$, 
we define a truncated modified energy of order $\eta^N$:
\begin{equation}
    \widehat{H}_\eta^{[N]}(z) := \sum_{r=0}^N\eta^r H_r(z) = H_0(z) + \sum_{r=1}^N\eta^rH_r(z).
    \label{eq:truncated-modified-energy}
\end{equation}
\begin{remark}
    \citet{katona2026symplectic} considered the same modified energy,
    then used a \emph{finite} truncation of the modified energy $\widehat{H}_\eta$ and proved its approximate preservation.
    In particular, they verified approximate preservation of the truncated modified energy up to order three in arbitrary dimension.
    However, this finite truncation still leaves an upper bound on $H_0(z_T) - H_0(z_0)$ with a polynomial dependence in $T$, 
    which leads to an $O(T^{1/5})$ regret bound.
    In contrast, we consider the truncated modified energy for arbitrary order $N$.
    We show that, for \emph{any} integer $N \geq 0$, 
    we can truncate the modified energy $\widehat{H}_\eta$ at order $N$, 
    and this truncated energy is always approximately preserved up to a sufficiently small error.
    We then take $N=\Theta(\log T)$ and use this $T$-dependent truncation to obtain an $O(\log T)$ regret bound.
\end{remark}

By utilizing the coefficient-by-coefficient equality up to $N$ order, and the upper bounds on the higher-order terms specified in~\cref{ass:regularity}, 
we can establish precise control over the truncated modified energy.
In particular, we establish the following bounds for the truncated modified energy, which is the central component of our analysis. 

\begin{lemma}
\label{lem:modified-energy-tool}
Under Assumption~\ref{ass:regularity}, set $C_0=24\max\{\norm{J_A}_{\op},1\}K^2$.
For every integer $N \geq 0$ and stepsize $0 < \eta \leq \frac{1}{16C_0(N+2)}$,
the truncated modified energy $\widehat{H}_\eta^{[N]}$ defined in~\cref{eq:truncated-modified-energy} satisfies,
uniformly over all dual coordinates $z$,
\begin{equation}
    \begin{aligned}
        \left|\widehat H_\eta^{[N]}(z)-H_0(z)\right|
        &\leq \frac{64}{3}C_0^2\eta, \\
        \left|\widehat H_\eta^{[N]}(\mathcal M_\eta(z))
        -\widehat H_\eta^{[N]}(z)\right|
        &\leq 16e^2C_0^2\eta\,4^{-(N+1)}.
    \end{aligned}
    \label{eq:modified-energy-tool}
\end{equation}
\end{lemma}
The first bound in~\cref{lem:modified-energy-tool} controls 
the deviation of the truncated modified energy $\widehat{H}_\eta^{[N]}$ from the original energy $H_0$.
The second bound quantifies the one-step near-preservation of $\widehat{H}_\eta^{[N]}$.
It is crucial to guarantee the \emph{exponential decay} of the one-step gap in the truncated modified energy as $N$ increases,
when we can take $\eta = \Theta(1/N)$.

With these tools from~\cref{eq:modified-energy-tool} in hand, 
we can control the change in the original energy $H_0$
along the trajectory of the AltOMD updates.
In particular, we can then bound $H_0(z_T)-H_0(z_0)$ by
\begin{align*}
    &\big(H_0(z_T)-\widehat H_\eta^{[N]}(z_T)\big) 
    + \widehat H_\eta^{[N]}(z_T)-\widehat H_\eta^{[N]}(z_0) 
    - \big(H_0(z_0)-\widehat H_\eta^{[N]}(z_0)\big) \\
    \leq& \left|H_0(z_T)-\widehat H_\eta^{[N]}(z_T)\right|
    + \sum_{t=0}^{T-1} \left|\widehat H_\eta^{[N]}(z_{t+1})-\widehat H_\eta^{[N]}(z_t)\right|
    + \left|H_0(z_0)-\widehat H_\eta^{[N]}(z_0)\right| \\
    \leq& 16e^2C_0^2 \eta T 4^{-(N+1)} + \frac{128}{3}C_0^2\eta.
\end{align*}

By~\cref{lem:regret-decomposition}, when $\eta = \Theta(1/N)$, the regret can be bounded by
\begin{equation*}
    \begin{aligned}
        \Reg_T
        &\leq \frac{1}{\eta} \left( 16e^2C_0^2\eta T4^{-(N+1)} + \frac{128}{3}C_0^2\eta + \Lambda_0 \right)\\
        &=16e^2C_0^2T4^{-(N+1)}+\frac{128}{3}C_0^2+\frac{\Lambda_0}{\eta}.
    \end{aligned}
\end{equation*}
Therefore, we can set $N = \Theta(\log{T})$ and obtain an $O(\log T)$ regret bound.
For integer $T\geq1$ and $N=\lfloor\log_2T\rfloor$, we have $T4^{-(N+1)} \leq \frac{1}{4}$:
this is equality for $T=1$, while for $T\geq2$, $T<2^{N+1}\leq4^N$.
\begin{theorem}[Logarithmic regret of AltOMD]
    \label{thm:main}
    Under Assumptions~\ref{ass:regularity}--\ref{ass:radius},
    for every integer time horizon $T\geq 1$,
    set $N = \lfloor \log_2 T\rfloor$ and $\eta = \frac{1}{16C_0(N+2)}$. 
    Let $(x_t, y_t)_{t=0}^T$ be the sequence of iterates generated by the AltOMD updates with stepsize $\eta$.
    Then, the regret defined in~\cref{eq:regret} satisfies 
    \begin{equation}
        \begin{aligned}
            \Reg_T &\leq 16C_0\Lambda_0(N+2)+\left(\frac{128}{3}+4e^2\right)C_0^2\\
            &\leq 16C_0\Lambda_0(\log_2 T+2)+\left(\frac{128}{3}+4e^2\right)C_0^2.
        \end{aligned}
        \label{eq:rate}
    \end{equation}
\end{theorem}

\cref{thm:main} applies to arbitrary bilinear games with any compact convex strategy sets 
satisfying the stated assumptions, including games with non-interior or non-unique equilibria.
By~\cref{eq:regret-gap}, the regret bound also implies
$\Gap(\bar x_T,\bar y_T)=O(\log T/T)$.
To the best of our knowledge, this is the first $O(\log T)$ regret bound for alternating online mirror descent algorithm \emph{with constant stepsize} in arbitrary bilinear games.
This proves that, surprisingly, with only the alternation, the OMD algorithm achieves logarithmic regret in arbitrary bilinear games.

\section{AltOMD with the Negative Entropy Regularizers}
\label{sec:examples}
In this section, we apply the logarithmic regret bound in~\cref{thm:main} to negative entropy regularizers on the probability simplex strategy sets, 
which is the well-known multiplicative weights update algorithm. 
We first describe the regularizers and the resulting AltOMD updates, then verify the assumptions of the theorem.

Let $\Delta_d := \{p \in \R^d: p_i \geq 0, \sum_{i=1}^d p_i = 1 \}$,
and set $\mathcal{X}=\Delta_n$ and $\mathcal{Y}=\Delta_m$.
For each player, use the negative entropy regularizer 
\begin{equation}
    \psi_d(p):=\sum\nolimits_{i=1}^d p_i\log p_i,\qquad p\in\Delta_d,
    \label{eq:shannon-mirror}
\end{equation}
with the convention $0\log0=0$ and $\psi_d(p)=+\infty$ outside $\Delta_d$.
That is, we set $\psi_{\mathcal{X}}=\psi_n$ and $\psi_{\mathcal{Y}}=\psi_m$.
The convex conjugates of the negative entropy regularizers are log-sum-exp functions, 
and their gradients are the softmax maps: 
\begin{equation}
    \begin{aligned}
        F(u)&=\log\sum\nolimits_{i=1}^n e^{u_i},
        & [\nabla F(u)]_i&=\frac{e^{u_i}}{\sum_{\ell=1}^n e^{u_\ell}},\\
        G(w)&=\log\sum\nolimits_{j=1}^m e^{w_j},
        & [\nabla G(w)]_j&=\frac{e^{w_j}}{\sum_{\ell=1}^m e^{w_\ell}}.
    \end{aligned}
    \label{eq:entropy-mirror-map}
\end{equation}

We substitute the softmax maps into~\eqref{eq:altomd-dual} and obtain the multiplicative weights update algorithm: 
\begin{equation}
    \begin{aligned}
        x_{k+1,i} &\propto x_{k,i}\exp(-\eta(A^\top y_k)_i) \quad \forall\, i=1, \ldots, n, \\
        y_{k+1,j} &\propto y_{k,j}\exp(\eta(Ax_{k+1})_j) \quad \forall\, j=1, \ldots, m, 
    \end{aligned}
    \label{eq:entropy}
\end{equation}
where $x_{k, i}$ and $y_{k, j}$ are the $i$-th and $j$-th components of the strategies at iteration $k$, respectively.

The following proposition verifies the assumptions of~\cref{thm:main} for the negative entropy regularizers.
\begin{proposition}[Validity of negative entropy regularizers]
    \label{prop:entropy-assumptions}
    If both players use negative entropy regularizers, i.e, $\psi_{\mathcal{X}} = \psi_n$ and $\psi_{\mathcal{Y}} = \psi_m$, then 
    the regularizers are Legendre relative to the affine hull of their respective simplices.
    Moreover, their convex conjugates $F$ and $G$ are smooth on $\R^n$ and $\R^m$, respectively,  
    and for every integer $s\geq1$,
    \begin{equation*}
        \frac{1}{s!}\left(
        \sup_u\norm{D^sF(u)}_{\op}+\sup_w\norm{D^sG(w)}_{\op}\right)
        \leq 4^s.
    \end{equation*}
    For any strictly positive initial strategies $x_0\in\Delta_n$ and $y_0\in\Delta_m$,
    the initial Bregman divergence satisfies
    \begin{equation*}
        \Lambda_0\leq\log\frac{1}{\min_{1\leq i\leq n}x_{0,i}}
        +\log\frac{1}{\min_{1\leq j\leq m}y_{0,j}}<\infty.
    \end{equation*}
    Thus, Assumptions~\ref{ass:regularity}--\ref{ass:radius} hold with $K=4$.
\end{proposition}
The proofs in this section are provided in~\cref{app:shannon}.
We apply~\cref{thm:main} to obtain the following corollary.
\begin{corollary}[Alternating multiplicative weights update]\label{cor:entropy}
    Let $A\in\R^{m\times n}$ and let $x_0\in\Delta_n$ and $y_0\in\Delta_m$
    be strictly positive initial strategies. 
    Set 
    $C = 384\max\{\norm{A}, 1\}$
    and $\Lambda_0 = \log({1}/{\min_{1\leq i\leq n}x_{0,i}}) + \log({1}/{\min_{1\leq j\leq m}y_{0,j}})$.
    For every integer $T\geq1$,
    choose the constant step size $\eta = \frac{1}{16C(\lfloor\log_2T\rfloor+2)}$,
    and generate a sequence of iterates for $k=0,\ldots,T-1$ by the updates in~\cref{eq:entropy}.
    Then, the total regret satisfies
    $\Reg_T \leq 16 C \Lambda_0(\log_2T+2)+\left(\frac{128}{3}+4e^2\right)C^2$.
    In particular, the regret bound for uniform initial strategies is
    \begin{equation}
        \Reg_T \leq 16C\log(nm)(\log_2T+2)+\left(\frac{128}{3}+4e^2\right)C^2.
        \label{eq:entropy-regret}
    \end{equation}
\end{corollary}
This corollary shows that the alternating multiplicative weights update algorithm achieves an $O(\| A \| \log(nm)\log(T) + \| A \|^2)$ regret bound.
By~\cref{eq:regret-gap}, the averaged strategies satisfy
$\Gap(\bar x_T,\bar y_T) = \widetilde{O}(\log T/T)$.

\section*{Acknowledgments}
The research of Christian Kroer was supported by the Office of Naval Research awards N00014-23-1-2374 and N00014-26-1-2490, and the National Science Foundation award IIS-2238960.

\section*{LLM usage}
The mathematical proof in this paper was developed using GPT-5.6 Sol.
The Lean proof was formalized using GPT-6 Astra.
The authors verified the proofs and revised the manuscript throughout to improve readability and exposition.

\bibliographystyle{plainnat}
\bibliography{references}

\appendix

\newpage

\begin{center}
    {\LARGE \textbf{Appendices}}
\end{center}

\section{Full proof of the main theorem}
\label{app:proof}

\begin{mainrestatement}[Logarithmic regret of AltOMD]
    Under Assumptions~\ref{ass:regularity}--\ref{ass:radius},
    for every integer time horizon $T\geq 1$,
    set $N = \lfloor \log_2 T\rfloor$ and $\eta = \frac{1}{16C_0(N+2)}$.
    Let $(x_t, y_t)_{t=0}^T$ be the sequence of iterates generated by the AltOMD updates with stepsize $\eta$.
    Then, the regret defined in~\cref{eq:regret} satisfies
    \begin{equation*}
        \begin{aligned}
            \Reg_T &\leq 16C_0\Lambda_0(N+2)+\left(\frac{128}{3}+4e^2\right)C_0^2\\
            &\leq 16C_0\Lambda_0(\log_2 T+2)+\left(\frac{128}{3}+4e^2\right)C_0^2.
        \end{aligned}
    \end{equation*}
\end{mainrestatement}

Before starting the proofs, we first explain some notations here. We extend the conjugates $F$ and $G$ to the product dual space by
$F(z)=F(u)$ and $G(z)=G(w)$, where $z=(u,w)\in\R^n\times\R^m$.
For a smooth Hamiltonian (energy) $Q$, 
its Hamiltonian flow $\phi_Q^t$ is defined, whenever the solution exists, by
\begin{equation}
    \frac{d}{dt}\phi_Q^t(z)=J_A\nabla Q(\phi_Q^t(z)),\qquad
    \phi_Q^0(z) = z,
    \label{eq:hamiltonian-flow}
\end{equation}
where $J_A$ is defined in~\cref{eq:poisson-matrix}.
Recall the one-step AltOMD map from the main text:
\begin{equation*}
 \mathcal M_\eta=\phi_F^\eta\circ\phi_G^\eta.
\end{equation*}

\subsection{Formal preservation and BCH correction bounds}
\label{app:bch}

In this subsection, we construct the BCH correction terms $H_r$ and prove formal preservation.
We bound their coefficients for the derivatives in the next subsection.
Throughout this subsection, we consider formal power series: 
series identities hold coefficient by coefficient, without assuming convergence.
Throughout the appendix, we write $\operatorname{set}(a,b)$ for the set whose elements are $a$ and $b$,
to distinguish set notation from the Poisson bracket.

\paragraph{Poisson bracket.}
For two smooth functions $P, Q$ on the product dual space, 
their \emph{Poisson bracket} is defined by
\begin{equation}
    \{P,Q\} := \nabla P^\top J_A\nabla Q.
    \label{eq:poisson-bracket}
\end{equation}
This operation is linear in each argument.
Because $J_A^\top=-J_A$, we have $\{P,Q\} = -\{Q,P\}$.

\paragraph{Jacobi identity.}
For smooth functions $P,Q,S$, the Poisson bracket satisfies
the Jacobi identity:
\begin{equation}
    \{P,\{Q,S\}\}+\{Q,\{S,P\}\}+\{S,\{P,Q\}\}=0.
    \label{eq:jacobi-identity}
\end{equation}
\begin{proof}
Since $J_A$ is constant and $J_A^\top=-J_A$, the product rule yields
\begin{align*}
    \{P,\{Q,S\}\}
    =-D\{Q,S\}[J_A\nabla P] = D^2S[J_A\nabla P,J_A\nabla Q]
    -D^2Q[J_A\nabla P,J_A\nabla S],
\end{align*}
with all derivatives evaluated at the same point.
In the cyclic sum of this identity over $P,Q,S$, all the terms cancel because
$D^2P$, $D^2Q$, and $D^2S$ are symmetric tensors.
This proves~\cref{eq:jacobi-identity}.
\end{proof}

\paragraph{Lie derivatives and update flows.}
For a Hamiltonian $Q$ and a smooth function $f$, define
\begin{equation}
    L_Q f := \{f,Q\}.
    \label{eq:lie-derivative}
\end{equation}
For any point $z$ in the product dual space, 
the chain rule yields
\begin{equation*}
    \frac{d}{dt}f(\phi_Q^t(z)) \overset{\textnormal{by~\cref{eq:hamiltonian-flow}}}{=} \nabla f(\phi_Q^t(z))^\top J_A \nabla Q(\phi^t_Q(z)) \overset{\textnormal{by~\cref{eq:lie-derivative}}}{=} (L_Qf)(\phi_Q^t(z)),
\end{equation*}
so $L_Q$ differentiates $f$ along the flow generated by $Q$.
In particular, $L_G$ and $L_F$ differentiate along the first and second
AltOMD update flows, respectively.
For $Q\in\operatorname{set}(F,G)$, repeated differentiation gives
\begin{equation*}
    e^{\eta L_Q}f := \sum_{k\geq0}\frac{\eta^k}{k!}L_Q^kf
    = f\circ\phi_Q^\eta,
\end{equation*}
where the right-hand side is interpreted as the formal Taylor series of the
composition at $\eta=0$, without assuming that this series converges.
Since $\mathcal M_\eta=\phi_F^\eta\circ\phi_G^\eta$, it follows that
\begin{equation}
    f\circ\mathcal M_\eta = f\circ\phi_F^\eta\circ\phi_G^\eta = e^{\eta L_G}e^{\eta L_F}f.
    \label{eq:formal-update-operator}
\end{equation}
More explicitly, Taylor expansion in independent parameters $s,t$ gives the formal identity
\begin{equation*}
    f\circ\phi_F^s\circ\phi_G^t
    =\sum_{\alpha,\beta\geq0}\frac{t^\alpha s^\beta}{\alpha!\beta!}
      L_G^\alpha L_F^\beta f.
\end{equation*}
Setting $s=t=\eta$ groups terms by total degree, with finitely many terms
in each coefficient. This explains both the reversed operator order and
the common parameter in~\cref{eq:formal-update-operator};
\cref{lem:real-taylor-composition} proves the corresponding finite Taylor formula with a remainder.
This identity applies to formal series of smooth functions by applying each operator to every coefficient.

\paragraph{BCH theorem.}
We first state the classical BCH theorem in universal form,
then apply it to the Hamiltonian Lie derivatives.
The accompanying Lean proof of the following theorem follows~\citet[Sections~1--2]{ha2026variation} and~\citet[Section~3]{hofstatter2021relatively}.

\begin{proposition}[BCH Theorem]
    \label{prop:bch}
    Let $U, V$ be free noncommuting symbols over $\mathbb Q$, i.e., we form
    polynomials in $U,V$ with rational coefficients, and the order of the symbols matters.
    Define the commutator $[P,Q]:=PQ-QP$.
    Then, there is a unique formal series $Z$ with zero constant term such that
    \begin{equation}
        e^Z = e^U e^V, 
        \label{eq:universal-bch-identity}
    \end{equation}
    where $Z$ is given by
    \begin{equation*}
        Z = \log(e^Ue^V) = \sum\nolimits_{q\geq1} \Phi_q(U,V),
    \end{equation*}
    where each $\Phi_q(U,V)$ is a homogeneous Lie polynomial of degree $q$:
    a rational linear combination of nested commutators containing $q$ factors
    from $U,V$~\citep{campbell1896law,campbell1897law,baker1905alternants,hausdorff1906symbolische,jacobson2013lie}.
    The explicit formula is given by Dynkin's formula~\citep{dynkin1947calculation}: for every $q \geq 1$, 
    \begin{equation}
        \Phi_q(U,V)
        =\frac1q\sum_{m=1}^{q}\frac{(-1)^{m-1}}{m}
        \sum_{\substack{r_i,s_i \geq 0,\, r_i+s_i>0\ (1\leq i\leq m)\\
            \sum_{i=1}^{m}(r_i+s_i)=q}}
        \frac{[U^{r_1}V^{s_1}\cdots U^{r_m}V^{s_m}]}
        {r_1!s_1!\cdots r_m!s_m!}, 
        \label{eq:dynkin-bch-coefficients}
    \end{equation}
    where $[W_1] := W_1$ and for $q \geq 2$, 
    \begin{equation*}
        [W_1\cdots W_q]
        :=[W_1,[W_2,\ldots,[W_{q-1},W_q]\ldots]],
        \qquad W_i\in\operatorname{set}(U,V).
    \end{equation*}
\end{proposition}

\paragraph{Application to the AltOMD update.}
By~\cref{eq:formal-update-operator}, one AltOMD step acts on a smooth function
$f$ through $f\circ\mathcal M_\eta=e^{\eta L_G}e^{\eta L_F}f$.
Here, products of Lie derivatives denote composition: 
$L_G L_F f = L_G(L_F f)$.
We apply~\cref{prop:bch} by substituting
\begin{equation*}
    U \leftarrow \eta L_G,\qquad V \leftarrow \eta L_F,
\end{equation*}
where $\eta$ is a formal scalar parameter that commutes with the operators.
By homogeneity of $\Phi_q$, it follows that 
\begin{equation}
    e^{Z(\eta)}=e^{\eta L_G}e^{\eta L_F}.
    \label{eq:BCH-expansion-exponential-product-identity}
\end{equation}
where 
\begin{equation}
    Z(\eta) = \sum\nolimits_{q\geq1} \eta^q Z_q, 
    \qquad Z_q := \Phi_q(L_G, L_F), 
    \label{eq:bch-logarithm}
\end{equation}
where the explicit form of $\Phi_q$ is given by~\cref{eq:dynkin-bch-coefficients}: 
\begin{equation}
    Z_q = \Phi_q(L_G, L_F)
    = \frac{1}{q} \sum_{m=1}^{q}\frac{(-1)^{m-1}}{m}
    \sum_{\substack{r_i,s_i \geq 0,\, r_i+s_i>0\ (1\leq i\leq m)\\
        \sum_{i=1}^{m}(r_i+s_i)=q}}
    \frac{[(L_G)^{r_1}(L_F)^{s_1}\cdots (L_G)^{r_m}(L_F)^{s_m}]}
    {r_1!s_1!\cdots r_m!s_m!}, 
    \label{eq:dynkin-bch-operator-coefficients}
\end{equation}
For example, 
\begin{equation*}
    \begin{aligned}
        Z_1 &= [L_G] + [L_F] = L_G + L_F,\\
        Z_2 &= \underbrace{\frac12\left(
            \frac{[L_G L_G]}{2!}+[L_G L_F]+\frac{[L_F L_F]}{2!}
            \right)}_{m=1} - \underbrace{\frac14\left(
            [L_G L_G]+[L_G L_F]+[L_F L_G]+[L_F L_F]
            \right)}_{m=2}\\
        &= \frac14[L_G L_F] - \frac14[L_F L_G] = \frac12[L_G,L_F].
    \end{aligned}
\end{equation*}
Here, we use $[L_G L_G] = [L_G, L_G] = 0$ and $[L_F L_F] = [L_F, L_F] = 0$.

Equivalently, 
\begin{align}
    Z_q 
    &= \frac{1}{q} \sum_{\omega=(Q_1,\ldots,Q_q)\in\operatorname{set}(G,F)^q}
    c_\omega[L_{Q_1},[L_{Q_2},\ldots,[L_{Q_{q-1}},L_{Q_q}]\ldots]], 
    \label{eq:bch-commutator-coefficients}
\end{align}
where 
\begin{equation}
    c_\omega
    =\sum_{m=1}^{q}\frac{(-1)^{m-1}}m
    \sum_{\substack{r_i, s_i \geq 0, \, r_i+s_i>0 \, (1\leq i\leq m)\\
        G^{r_1}F^{s_1}\cdots G^{r_m}F^{s_m}=\omega}}
    \frac1{r_1!s_1!\cdots r_m!s_m!}.
    \label{eq:bch-scalar-coefficients}
\end{equation}

\paragraph{Complete construction of the modified Hamiltonian (energy).}

To construct the modified Hamiltonian energy in~\cref{eq:formal-modified-energy-sketch}, 
we construct each correction term $H_{q-1}$ corresponding to the operator $Z_q$ in~\cref{eq:bch-commutator-coefficients}.

Recall that, for two operators $U,V$, the commutator is defined by $[U,V] = UV - VU$.
By the Jacobi identity, we have
\begin{align}
    [L_P,L_Q]f
    =\{\{f,Q\},P\}-\{\{f,P\},Q\} =\{f,\{Q,P\}\} = \{f, -\{P, Q\} \} = - L_{\{P, Q\}} f.
    \label{eq:opposite-bracket}
\end{align}
For any $Q_1,\ldots,Q_q\in\operatorname{set}(G,F)$, this relation implies by induction that
\begin{equation}
    [L_{Q_1},[L_{Q_2},\ldots,[L_{Q_{q-1}},L_{Q_q}]\ldots]]
    =(-1)^{q-1}L_{\{Q_1,\{Q_2,\ldots,\{Q_{q-1},Q_q\}\ldots\}\}}.
    \label{eq:nested-hamiltonian-operators}
\end{equation}

For every integer $q\geq1$, we define the correction terms by
\begin{equation}
    H_{q-1}:=\frac{(-1)^{q-1}}q
    \sum_{\omega=(Q_1,\ldots,Q_q)\in\operatorname{set}(G,F)^q}
    c_\omega\{Q_1,\{Q_2,\ldots,\{Q_{q-1},Q_q\}\ldots\}\},
    \label{eq:hamiltonian-corrections}
\end{equation}
where $c_\omega$ are defined by~\cref{eq:bch-scalar-coefficients}.
For $q=1$, the nested bracket is interpreted as $Q_1$ thus $H_0=F+G$.
By~\cref{eq:bch-commutator-coefficients,eq:nested-hamiltonian-operators} and the fact that 
the $L_Q$ is linear in $Q$: 
\begin{equation*}
    L_{\sum_i a_i Q_i} f = \left\{ f, \sum_i a_i Q_i \right\} = \sum_i a_i \{f, Q_i\} = \left( \sum_i a_i L_{Q_i} \right) f, 
\end{equation*}
we obtain 
\begin{equation}
    Z_q = L_{H_{q-1}}, \quad \forall \ q \geq 1.
    \label{eq:bch-hamiltonian-operators}
\end{equation}
The formal modified Hamiltonian energy is therefore 
\begin{equation}
    \widehat H_\eta:=\sum_{r\geq0}\eta^rH_r. 
    \label{eq:formal-modified-hamiltonian}
\end{equation}
Again, because $L_Q$ is linear in $Q$, we have 
\begin{equation}
    Z(\eta) = \eta L_{\widehat H_\eta}.
    \label{eq:formal-Z-operator}
\end{equation}
For example, 
\begin{equation}
    \begin{aligned}
        H_0 &= F + G, \quad Z_1 = L_{H_0} \\
        H_1 &= \frac{1}{2} \{F, G\}, \quad Z_2 = L_{H_1}.
    \end{aligned}
    \label{eq:first-shadow-term}
\end{equation}

\paragraph{Formal preservation of the modified energy.}

\begin{formalpreservationrestatement}[Formal preservation of $\widehat{H}_\eta$]
    Let $F$ and $G$ be smooth, let $H_r$ be the BCH correction terms constructed in~\cref{app:bch}, and let $\widehat H_\eta$ be the formal modified energy defined by~\cref{eq:formal-modified-energy-sketch}.
    Then, $\widehat H_\eta$ satisfies
    \begin{equation*}
        \widehat H_\eta\bigl(\mathcal M_\eta(z)\bigr) = \widehat H_\eta(z)
        \qquad\text{for every } z \in \R^n \times \R^m,
\end{equation*}
    as an identity of formal power series in $\eta$.
\end{formalpreservationrestatement}

\begin{proof}[Proof of~\cref{lem:formal-modified-energy-preservation}]
By~\cref{eq:formal-Z-operator}, we have $Z(\eta) = \eta L_{\widehat H_\eta}$.
By BCH Theorem,~\cref{eq:BCH-expansion-exponential-product-identity} yields
\begin{equation*}
    e^{\eta L_G}e^{\eta L_F} = e^{\eta L_{\widehat H_\eta}}.
\end{equation*}
By antisymmetry of the Poisson bracket,
\begin{equation*}
    L_{\widehat H_\eta}\widehat H_\eta
    =\{\widehat H_\eta,\widehat H_\eta\} 
    = (\nabla \widehat H_\eta)^\top J_A (\nabla \widehat H_\eta) 
    = \frac{1}{2} (\nabla \widehat H_\eta)^\top J_A (\nabla \widehat H_\eta) + \frac{1}{2} (\nabla \widehat H_\eta)^\top J_A^\top (\nabla \widehat H_\eta) 
    = 0.
\end{equation*}
Therefore,
\begin{equation}
    e^{\eta L_G}e^{\eta L_F}\widehat H_\eta=e^{\eta L_{\widehat H_\eta}}\widehat H_\eta=\widehat H_\eta+\sum_{k\geq1}\frac{\eta^k}{k!}L_{\widehat H_\eta}^{\,k}\widehat H_\eta=\widehat H_\eta, 
    \label{eq:formal-scalar-invariance}
\end{equation}
where the second equality uses the formal exponential series 
$e^{\eta L_{\widehat H_\eta}}=\sum_{k\geq0}(\eta L_{\widehat H_\eta})^k/k!$.
By~\cref{eq:formal-update-operator,eq:formal-scalar-invariance}, we have
$\widehat H_\eta\circ\mathcal M_\eta=\widehat H_\eta$, which proves the lemma.
\end{proof}

\paragraph{Coefficient bounds for the correction terms.}
The derivative estimates in the next subsection require a bound on the
coefficients in a representation of each $H_r$ by nested Poisson brackets.
The formula~\cref{eq:hamiltonian-corrections} reduces this task to a bound on
$\frac1q\sum_\omega|c_\omega|$, where $q=r+1$.
We estimate this sum directly from the formal logarithm.

For a finite linear combination $P$ of distinct labelled operator words,
let $a_\omega$ denote the coefficient of the word $\omega$. Define
the norm $\|P\|_1:=\sum_\omega|a_\omega|$, the sum of the
absolute values of the coefficients $a_\omega$.  In particular,
$\|Z_q\|_1=\sum_{\omega\in\operatorname{set}(G,F)^q}|c_\omega|$.
This is a norm on formal word expressions, not a norm on the evaluated
differential operators; the empty word represents the identity operator $I$.
Let $[\eta^q]f(\eta)$ denote the coefficient of $\eta^q$ in a formal series $f(\eta)$.
We use the same coefficient notation for series in other variables,
including those with function or operator coefficients.
Set
\begin{equation*}
    E(\eta) := e^{\eta L_G}e^{\eta L_F} - I
    = \sum_{\ell \geq 1} E_\ell \eta^\ell,
    \qquad
    E_\ell
    := \sum_{j=0}^{\ell}
    \frac{L_G^j L_F^{\ell-j}}{j! (\ell-j)!}.
\end{equation*}
All coefficients of $E_\ell$ are nonnegative, and the binomial theorem yields
\begin{equation}
    \|E_\ell\|_1
    =\sum_{j=0}^{\ell}\frac{1}{j!(\ell-j)!}
    =\frac{2^\ell}{\ell!} \leq 2^\ell.
    \label{eq:E-norm}
\end{equation}

For any two such polynomials $P$ and $Q$, this norm satisfies
\begin{equation*}
    \|PQ\|_1 \leq \|P\|_1\|Q\|_1.
\end{equation*}
To see why, index the coefficients of $P$ and $Q$ by labelled words
$\sigma$ and $\tau$, respectively.
The product of two words is the concatenation of their lists of Lie derivatives.
The coefficient contributed by this product is $a_\sigma b_\tau$,
and the sum of the absolute values of their coefficients is
$\sum_{\sigma,\tau}|a_\sigma b_\tau|=(\sum_\sigma|a_\sigma|)(\sum_\tau|b_\tau|)$.
By the triangle inequality, this sum cannot increase when we collect terms with the same word.
The formal logarithm has the expansion
\begin{equation*}
    Z(\eta)=\log(I+E(\eta))=\sum\nolimits_{k\geq1}\frac{(-1)^{k+1}}{k}E(\eta)^k.
\end{equation*}
The powers $E(\eta)^k$ contribute to the coefficient of $\eta^q$ as follows:
\begin{equation*}
    Z_q = \sum_{k=1}^{q}\frac{(-1)^{k+1}}{k}
    \sum_{\substack{\ell_1+\cdots+\ell_k=q\\ \ell_i\geq1}}
    E_{\ell_1}\cdots E_{\ell_k}.
\end{equation*}
The triangle inequality and the bound $\|PQ\|_1\leq\|P\|_1\|Q\|_1$ imply
\begin{align}
    \|Z_q\|_1
    \leq \sum_{k=1}^{q}\frac{1}{k}
    \sum_{\substack{\ell_1+\cdots+\ell_k=q\\ \ell_i\geq1}}
    \|E_{\ell_1}\|_1\cdots\|E_{\ell_k}\|_1
    &= [\eta^q]\sum_{k\geq1}\frac{1}{k}
    \left(\sum_{\ell\geq1}\|E_\ell\|_1\eta^\ell\right)^k
    \notag\\
    &\leq [\eta^q] \sum_{k \geq 1} \frac{1}{k}
    \left(\sum_{\ell \geq 1} 2^\ell \eta^\ell\right)^k
    \tag{by~\cref{eq:E-norm}} \\
    &= [\eta^q] \log\left(\frac{1 - 2\eta}{1 - 4\eta}\right)
    =
    \frac{4^q - 2^q}{q}
    \leq 4^q,
    \label{eq:bch-mass}
\end{align}
where the last line follows from $\sum_{\ell\geq1}2^\ell \eta^\ell=2\eta/(1-2\eta)$
and $-\log(1-\eta)=\sum_{k\geq1}\eta^k/k$.

Together,~\cref{eq:bch-commutator-coefficients,eq:bch-mass} bound the coefficients in the
right-nested commutator representation of $Z_q$.  Before like terms are
collected, the sum of the absolute values of the coefficients in this representation of $Z_q$ is at most
\begin{equation}
 \frac{1}{q}\sum_{\omega\in\operatorname{set}(G,F)^q}|c_\omega|
 =\frac{1}{q}\|Z_q\|_1\leq 4^q.
\label{eq:bch-bracket-mass}
\end{equation}

\begin{lemma}[Coefficient bound for the correction terms]
\label{lem:correction-coefficients}
For every integer $r\geq0$, $H_r$ can be written as a linear combination of
right-nested Poisson brackets of length $r+1$ in $F,G$, with the sum of the
absolute values of the coefficients at most $4^{r+1}$.
For $r=0$, length-one brackets mean $F$ and $G$ themselves.
\end{lemma}

\begin{proof}
With $q=r+1$, \cref{eq:hamiltonian-corrections} is already a representation
of $H_r$ by right-nested Poisson brackets of length $r+1$.
The common sign $(-1)^{q-1}$ does not affect the absolute coefficient sum,
which is $\frac1q\sum_{\omega\in\operatorname{set}(G,F)^q}|c_\omega|$
and is at most $4^q=4^{r+1}$ by~\cref{eq:bch-bracket-mass}.
For $r=0$, the length-one convention gives $H_0=F+G$.
\end{proof}

\paragraph{Truncation and formal energy error.}
For every integer $N \geq 0$, recall from the main text that the truncated modified energy is 
$\widehat H_\eta^{[N]}=\sum_{r=0}^N\eta^rH_r$.
In the subsequent analysis, for every integer $k\geq0$, $O(\eta^k)$ denotes a formal series of functions or operators whose coefficients below degree $k$ all vanish.
The omitted terms satisfy 
\begin{equation*}
    \widehat H_\eta-\widehat H_\eta^{[N]} = \sum\nolimits_{r\geq N+1}\eta^r H_r = O(\eta^{N+1}).
\end{equation*}
Since $e^{\eta L_G}e^{\eta L_F} - I = (I + \eta L_G + O(\eta^2))(I + \eta L_F + O(\eta^2)) - I = O(\eta)$, we have 
\begin{align}
    \widehat H_\eta^{[N]}\circ\mathcal M_\eta-\widehat H_\eta^{[N]}
    &=(e^{\eta L_G}e^{\eta L_F}-I)\widehat H_\eta^{[N]} \notag\\
    &=(e^{\eta L_G}e^{\eta L_F}-I) \left( \widehat{H}_\eta - \sum\nolimits_{r\geq N+1}\eta^r H_r \right) \notag\\
    &=-(e^{\eta L_G}e^{\eta L_F}-I) \sum\nolimits_{r\geq N+1}\eta^rH_r \tag{by~\cref{lem:formal-modified-energy-preservation}} \\
    &=O(\eta^{N+2}).
    \label{eq:finite-cancellation}
\end{align}
Therefore, every Taylor coefficient of the one-step energy change through degree $N+1$ is zero. 
In the next subsection, we use finite Taylor argument by combining this cancellation with the derivative bounds 
to control the actual change in $\widehat H_\eta^{[N]}$ over one iteration.

\subsection{Derivative bounds and finite modified-energy estimates}
\label{app:finite-energy}

We first bound the derivatives of nested Poisson brackets and
the correction terms $H_r$,
then use finite Taylor expansions to control the truncated modified energy.
For the estimates in this subsection, write $J=J_A$ and
$\|B\|=\|B\|_{\op}$ for the spectral norm of a matrix $B$.
For a smooth scalar- or vector-valued map $f$ on the product dual space, set
\begin{equation*}
    \|D^sf\|_\infty:=\sup_{z\in\R^n\times\R^m}\|D^sf(z)\|_{\op},
    \qquad s\geq1,
\end{equation*}
using Euclidean norms on the inputs and output.
For $s=0$, $\|f\|_\infty$ is the supremum of the output norm,
with absolute value for scalar-valued $f$.
In this notation,~\cref{ass:regularity} states that
$\|D^sF\|_\infty+\|D^sG\|_\infty\leq K^s s!$ for every integer $s\geq1$.
The Taylor formulas below use a variable stepsize $h$ and are applied with $h=\eta$.
We use the formal identities from~\cref{app:bch} to identify Taylor coefficients.

\begin{modifiedenergyrestatement}
Under Assumption~\ref{ass:regularity}, set $C_0=24\max\{\norm{J_A}_{\op},1\}K^2$.
For every integer $N \geq 0$ and stepsize $0 < \eta \leq \frac{1}{16C_0(N+2)}$,
the truncated modified energy $\widehat{H}_\eta^{[N]}$ defined in~\cref{eq:truncated-modified-energy} satisfies,
uniformly over all dual coordinates $z$,
\begin{equation*}
    \begin{aligned}
        \left|\widehat H_\eta^{[N]}(z)-H_0(z)\right|
        &\leq \frac{64}{3}C_0^2\eta, \\
        \left|\widehat H_\eta^{[N]}(\mathcal M_\eta(z))
        -\widehat H_\eta^{[N]}(z)\right|
        &\leq 16e^2C_0^2\eta\,4^{-(N+1)}.
    \end{aligned}
\end{equation*}
\end{modifiedenergyrestatement}
The functions $H_r$ and the constant $C_0$ are independent of $\eta$, $N$, and the horizon.

We first bound the derivatives of a single Lie derivative, then apply this bound repeatedly to nested Poisson brackets.
Recall that we use $[t^s]b(t)$ to denote the coefficient of $t^s$ in a formal scalar power series $b(t)$,
and use $\preceq$ to denote coefficient-wise inequality between such series.

\begin{lemma}[Coefficient bound for a Lie derivative]
\label{lem:lie-derivative-majorant}
Under~\cref{ass:regularity}, let $Q\in\operatorname{set}(F,G)$, let $f:\mathbb R^n\times\mathbb R^m\to\mathbb R$ be smooth, and set $v(t)=\|J\|K(1-Kt)^{-2}$.
If $b(t)$ is a formal power series with nonnegative coefficients satisfying
\begin{equation*}
    \frac{\|D^sf\|_\infty}{s!}\leq [t^s]b(t),\qquad \forall\,s\geq1,
\end{equation*}
then
\begin{equation*}
    \frac{\|D^s(L_Qf)\|_\infty}{s!}
    \leq [t^s]\bigl(v(t)b'(t)\bigr),\qquad \forall\,s\geq0.
\end{equation*}
\end{lemma}

\begin{proof}
Because $J$ is a constant matrix,
the derivatives of the Hamiltonian vector field satisfy
$\|D^\ell(J\nabla Q)\|_\infty \leq \|J\|\,\|D^{\ell}(\nabla Q)\|_\infty = \|J\|\,\|D^{\ell+1}Q\|_\infty$.
For every integer $\ell \geq 0$, \cref{ass:regularity} at order $\ell+1$ implies
\begin{equation}
    \frac{\|D^\ell(J\nabla Q)\|_\infty}{\ell!} \leq \frac{\|J\|\,\|D^{\ell+1}Q\|_\infty}{\ell!}
    \leq \|J\|\frac{K^{\ell+1}(\ell+1)!}{\ell!}
    = \|J\|K(\ell+1)K^\ell.
    \label{eq:J-nabla-Q-derivative-bound}
\end{equation}
The coefficients of $v(t)$ bound the normalized derivative norms of $J\nabla Q$:
\begin{equation}
    \sum\nolimits_{\ell \geq 0} \frac{\|D^\ell(J\nabla Q)\|_\infty}{\ell!} t^\ell \preceq \|J\|K\sum\nolimits_{\ell\geq0}(\ell+1)(Kt)^\ell
    = \frac{\|J\|K}{(1-Kt)^2} = v(t),
    \label{eq:scalar-majorant}
\end{equation}
where the first equality follows by differentiating the formal geometric-series identity
$\sum_{\ell\geq0}K^\ell t^{\ell+1}=t/(1-Kt)$:
\begin{equation*}
    \sum\nolimits_{\ell\geq0}(\ell+1)(Kt)^\ell
    = \frac{d}{dt}\left(\frac{t}{1-Kt}\right)
    = \frac{1}{(1-Kt)^2}.
\end{equation*}
Hence, \cref{eq:scalar-majorant} tells us that
\begin{equation}
    \frac{\|D^\ell(J\nabla Q)\|_\infty}{\ell!}
    \leq [t^\ell]v(t),\qquad \forall\,\ell\geq0.
    \label{eq:J-nabla-Q-derivative-coefficient-bound}
\end{equation}

Because $L_Qf = \{ f, Q \} = (J\nabla Q)\cdot\nabla f$ and $J$ is a constant matrix,
by the higher-order product rule, we have
\begin{equation}
    \|D^s(L_Qf)\|_\infty
    \leq \sum_{\ell=0}^s\binom{s}{\ell}
    \|D^{\ell}(J\nabla Q)\|_\infty\,
    \|D^{s-\ell+1}f\|_\infty.
    \label{eq:lie-derivative-product-rule}
\end{equation}
We divide both sides of~\cref{eq:lie-derivative-product-rule} by $s!$
and use $\binom{s}{\ell}/s!=1/(\ell!(s-\ell)!)$. For every integer $s\geq0$, we obtain
\begin{align}
    \frac{\|D^s(L_Qf)\|_\infty}{s!}
    &\leq \sum_{\ell=0}^s
    \frac{\|D^{\ell}(J\nabla Q)\|_\infty}{\ell!}
    \frac{\|D^{s-\ell+1}f\|_\infty}{(s-\ell)!} \notag\\
    &\leq \sum_{\ell=0}^s
    \bigl([t^\ell]v(t)\bigr)
    (s-\ell+1)\bigl([t^{s-\ell+1}]b(t)\bigr) \notag\\
    &= \sum_{\ell=0}^s
    \bigl([t^\ell]v(t)\bigr)
    \bigl([t^{s-\ell}]b'(t)\bigr) \notag\\
    &= [t^s]\bigl(v(t)b'(t)\bigr),
    \label{eq:lie-derivative-coefficient-bound}
\end{align}
where the second inequality uses~\cref{eq:J-nabla-Q-derivative-coefficient-bound} and the assumed
coefficient bounds on the derivatives of $f$,
the first equality follows from $[t^k]b'(t)=(k+1)[t^{k+1}]b(t)$ for every integer $k\geq0$,
and the last equality is because the coefficient of $t^s$ in the product $v(t)b'(t)$ is contributed by $[t^\ell]v(t)$ and $[t^{s-\ell}]b'(t)$ for some $\ell\in\{0,1,\ldots,s\}$.
\end{proof}

We now iterate~\cref{lem:lie-derivative-majorant} to bound the derivatives of nested Poisson brackets.
\begin{lemma}[Derivative bound for nested brackets]
\label{lem:bracket-majorant}
Assume~\cref{ass:regularity} holds.
Let $q \geq 2$ be an integer and $Q_1,\ldots,Q_q\in\operatorname{set}(F,G)$.
For every integer $s \geq 0$, we have
\begin{equation}
    \frac{1}{(q+s-1)!}\left\|D^s\{Q_1,\{Q_2,\ldots,\{Q_{q-1},Q_q\}\ldots\}\}\right\|_\infty
    \leq C_0^{q+s-1}.
    \label{eq:bracket-factorial-audited}
\end{equation}
For $q=1$ and $s \geq 1$, the derivative bounds are reduced to~\cref{ass:regularity}.
\end{lemma}
\begin{proof}
Let $v(t)$ be the series defined in~\cref{lem:lie-derivative-majorant}.
Define
\begin{equation}
    b_1(t)=(1-Kt)^{-1} = \sum\nolimits_{\ell\geq0}K^\ell t^{\ell}.
\end{equation}
By~\cref{ass:regularity}, for every $Q\in\operatorname{set}(F,G)$ and integer $s\geq1$,
\begin{equation*}
    \frac{\|D^sQ\|_\infty}{s!}\leq K^s = [t^s]b_1(t).
\end{equation*}
The derivative of $b_1(t)$ satisfies
\begin{align*}
    b_1'(t)
    =\frac{d}{dt}\left(1 + \sum\nolimits_{\ell\geq1}K^\ell t^\ell\right)
    =\sum\nolimits_{\ell\geq1}\ell K^\ell t^{\ell-1}
    =\sum\nolimits_{\ell\geq0}(\ell+1) K^{\ell+1} t^{\ell}.
\end{align*}
In particular, for every integer $s\geq0$,
\begin{equation}
    \frac{\|D^{s+1}Q\|_\infty}{s!}
    \leq (s+1)K^{s+1}=[t^s]b_1'(t),
\end{equation}
where the first inequality uses~\cref{ass:regularity}.
Since $\{Q,f\}=-L_Qf$, the nested bracket satisfies
\begin{equation*}
    \{Q_1,\{Q_2,\ldots,\{Q_{q-1},Q_q\}\ldots\}\}
    =(-1)^{q-1}L_{Q_1}\cdots L_{Q_{q-1}}Q_q.
\end{equation*}
We start with $f=Q_q$ and apply~\cref{lem:lie-derivative-majorant}
successively with $Q=Q_{q-1},\ldots,Q_1$.
Define the corresponding scalar series recursively by
\begin{equation*}
    b_{j+1}(t):=v(t)b_j'(t),\qquad j=1,\ldots,q-1,
\end{equation*}
which has the explicit form
\begin{align}
    b_q(t) = \left(v(t)\frac{d}{dt}\right)^{q-1}b_1(t) = \|J\|^{q-1}K^{2q-2}
    \left[\prod_{\ell=0}^{q-2}(3\ell+1)\right](1-Kt)^{-(3q-2)}.
    \label{eq:bq-explicit-form}
\end{align}
Indeed, if $b_j(t)=c_j(1-Kt)^{-(3j-2)}$, then
\begin{equation*}
    v(t)b_j'(t)=\|J\|K^2(3j-2)c_j(1-Kt)^{-(3j+1)}.
\end{equation*}
Starting from $c_1=1$, induction gives the prefactor in~\cref{eq:bq-explicit-form},
with the empty product interpreted as $1$.
We show that $b_q$ is the correct scalar series corresponding to the nested bracket by induction on $q$.
For $q=2$, we can apply~\cref{lem:lie-derivative-majorant} with
$f=Q_2$, $Q=Q_1$, and $b=b_1$.
Since $\{Q_1,Q_2\}=-L_{Q_1}Q_2$, for every integer $s\geq0$,
\begin{equation}
    \frac{\|D^s\{Q_1,Q_2\}\|_\infty}{s!}
    =\frac{\|D^s(L_{Q_1}Q_2)\|_\infty}{s!} \leq [t^s]\bigl(v(t)b_1'(t)\bigr)
    =[t^s]b_2(t).
\end{equation}
Now, let $q \geq 3$ and assume the bound holds for all brackets with $q-1$ factors.
Set $f=\{Q_2,\{Q_3,\ldots,\{Q_{q-1},Q_q\}\ldots\}\}$. The induction hypothesis states that
\begin{equation*}
    \frac{\|D^sf\|_\infty}{s!}\leq [t^s]b_{q-1}(t),\qquad \forall\,s\geq0.
\end{equation*}
For this $f$, \cref{lem:lie-derivative-majorant} with $Q=Q_1$
and $b=b_{q-1}$ implies, for every integer $s\geq0$,
\begin{align}
    \frac{\left\|D^s\{Q_1,\{Q_2,\ldots,\{Q_{q-1},Q_q\}\ldots\}\}\right\|_\infty}{s!}
    =\frac{\|D^s(L_{Q_1}f)\|_\infty}{s!} \leq [t^s]\bigl(v(t)b_{q-1}'(t)\bigr) =[t^s]b_q(t).
    \label{eq:lie-derivative-coefficient-bound-nested}
\end{align}
It remains to compute $[t^s]b_q(t)$. To do so, first note that
\begin{align*}
    [t^s](1-Kt)^{-(3q-2)}
    =\frac{1}{s!}\left.\frac{d^s}{dt^s}(1-Kt)^{-(3q-2)}\right|_{t=0}
    =K^s\frac{(3q+s-3)!}{s!\,(3q-3)!}
    =K^s\binom{3q+s-3}{s}.
\end{align*}
Also, since $3\ell+1\leq3(\ell+1)$ for every $\ell\geq0$, we have
$\prod_{\ell=0}^{q-2}(3\ell+1) \leq\prod_{\ell=0}^{q-2}3(\ell+1) =3^{q-1}(q-1)!$.
By~\cref{eq:bq-explicit-form} and the bounds above, we have
\begin{equation}
    [t^s] b_q(t) \leq \|J\|^{q-1}K^{2q-2}3^{q-1}(q-1)! K^s\binom{3q+s-3}{s}.
    \label{eq:bq-coefficient-bound}
\end{equation}

We use~\cref{eq:bq-coefficient-bound} in~\cref{eq:lie-derivative-coefficient-bound-nested} to obtain
\begin{equation}
    \left\|D^s\{Q_1,\{Q_2,\ldots,\{Q_{q-1},Q_q\}\ldots\}\}\right\|_\infty
    \leq \|J\|^{q-1}K^{2q-2+s}3^{q-1}(q-1)!\,s!
    \binom{3q+s-3}{s}.
    \label{eq:nested-bracket-coefficient-estimate}
\end{equation}
The binomial theorem yields
\begin{equation}
    \binom{3q+s-3}{s} \leq \sum\nolimits_{j=0}^{3q+s-3}\binom{3q+s-3}{j} = 2^{3q+s-3}.
    \label{eq:binomial-bound}
\end{equation}
Since $\binom{q+s-1}{s}\geq1$, we also have
\begin{equation}
    (q-1)!\,s! =\frac{(q+s-1)!}{\binom{q+s-1}{s}} \leq(q+s-1)!.
    \label{eq:factorial-bound}
\end{equation}
The bounds in~\cref{eq:binomial-bound,eq:factorial-bound} reduce~\cref{eq:nested-bracket-coefficient-estimate} to
\begin{align}
    \left\|D^s\{Q_1,\{Q_2,\ldots,\{Q_{q-1},Q_q\}\ldots\}\}\right\|_\infty
    &\leq\|J\|^{q-1}K^{2q-2+s}3^{q-1}2^{3q+s-3}(q+s-1)! \notag\\
    &=(24\|J\|K^2)^{q-1}(2K)^s(q+s-1)! \notag\\
    &\leq C_0^{q+s-1}(q+s-1)!,
    \label{eq:nested-bracket-constant-bound}
\end{align}
where the last inequality follows from $C_0=24\max\{\|J\|,1\}K^2$ and $K\geq1$, which give
$24\|J\|K^2\leq C_0$ and $2K\leq24K^2\leq C_0$.
\end{proof}

We use~\cref{lem:bracket-majorant}
to control the derivatives of the coefficients $H_r$ of the modified energy along the two mirror descent update flows.
The operators $L_G$ and $L_F$ differentiate along the first and second update flows, respectively.
\begin{lemma}[Lie derivative bound for the correction terms]
\label{lem:lie-derivative-bound}
Assume~\cref{lem:bracket-majorant} holds.
For all nonnegative integers $r,\alpha,\beta$ with $r+\alpha+\beta \geq 1$,
we have
\begin{equation}
    \|L_G^\alpha L_F^\beta H_r\|_\infty
    \leq (4C_0)^{r+\alpha+\beta+1}(r+\alpha+\beta)!.
    \label{eq:lie-derivative-bound-audited}
\end{equation}
\end{lemma}

\begin{proof}
By~\cref{lem:correction-coefficients}, for each $r\geq0$, there is a finite representation
\begin{equation}
    H_r=\sum\nolimits_{i\in I_r}c_iB_i,
    \qquad \sum\nolimits_{i\in I_r}|c_i|\leq4^{r+1},
    \label{eq:correction-coefficients-representation}
\end{equation}
where $I_r$ is a finite index set and each $B_i$ is a right-nested Poisson bracket of length $r+1$ in $F,G$.

We first describe the effect of the Lie derivatives on a single term $B_i$.
Since $L_Qf=\{f,Q\}=-\{Q,f\}$, the operator $L_Q$ adds $Q$ as an outer
bracket and changes the sign.
To keep track of the order, define $B_i^{(0)}=B_i$ and, for $1\leq j\leq\alpha+\beta$,
\begin{equation}
    B_i^{(j)}=
    \begin{cases}
        \{F,B_i^{(j-1)}\},&1\leq j\leq\beta,\\
        \{G,B_i^{(j-1)}\},&\beta<j\leq\alpha+\beta.
    \end{cases}
    \label{eq:Bi-j-definition}
\end{equation}
Hence, the first $\beta$ brackets correspond to $L_F^\beta$, and the next
$\alpha$ correspond to $L_G^\alpha$.
By~\cref{eq:Bi-j-definition}, we have
\begin{equation}
    L_G^\alpha L_F^\beta B_i = (-1)^{\alpha+\beta}B_i^{(\alpha+\beta)}
\end{equation}
Moreover, by~\cref{eq:correction-coefficients-representation}, we have
\begin{equation}
    L_G^\alpha L_F^\beta H_r = \sum\nolimits_{i\in I_r} (-1)^{\alpha+\beta} c_i B_i^{(\alpha+\beta)}.
\end{equation}
Because $B_i$ is the right-nested Poisson bracket of length $r+1$, each $B_i^{(\alpha+\beta)}$ has length $r+\alpha+\beta+1\geq2$.
By~\cref{lem:bracket-majorant} with $q=r+\alpha+\beta+1$ and $s=0$, each such bracket satisfies
\begin{equation}
    \|B_i^{(\alpha+\beta)}\|_\infty
    \leq C_0^{r+\alpha+\beta}(r+\alpha+\beta)!,\qquad i\in I_r.
\end{equation}
By the triangle inequality and~\cref{lem:correction-coefficients}, we obtain
\begin{equation}
    \|L_G^\alpha L_F^\beta H_r\|_\infty
    \leq \sum_{i\in I_r}|c_i|\,\|B_i^{(\alpha+\beta)}\|_\infty
    \leq4^{r+1}C_0^{r+\alpha+\beta}(r+\alpha+\beta)!
    \leq(4C_0)^{r+\alpha+\beta+1}(r+\alpha+\beta)!,
\end{equation}
where the last inequality is by $\alpha+\beta\geq0$ and $C_0\geq1$.
\end{proof}

We next bound the Taylor remainder of $f\circ\mathcal M_h$,
then apply this bound to the truncated modified energy $\widehat H_h^{[N]}$.

\begin{lemma} %
\label{lem:real-taylor-composition}
Fix an integer $m\geq0$ and a smooth scalar function $f$.
If
\begin{equation}
 \|L_G^\alpha L_F^\beta f\|_\infty\le M
 \quad\text{whenever }\alpha+\beta=m+1,
\end{equation}
then
\begin{equation}
 \left|f(\mathcal M_hz)-
 \sum_{\alpha+\beta\le m}
 \frac{h^{\alpha+\beta}}{\alpha!\beta!}
 L_G^\alpha L_F^\beta f(z)\right|
 \le \frac{2^{m+1}}{(m+1)!}Mh^{m+1},\qquad h\ge0.
\label{eq:composed-taylor-remainder}
\end{equation}
\end{lemma}

\begin{proof}
Fix a point $z$ in the dual space and a stepsize $h\geq0$.
For $Q\in\operatorname{set}(F,G)$, the flow satisfies $\frac{d}{dt}\phi_Q^t(z)=J\nabla Q(\phi_Q^t(z))$.
For any smooth scalar function $g$, the chain rule therefore yields
\begin{equation}
    \frac{d}{dt}g(\phi_Q^t(z)) = \nabla g(\phi_Q^t(z))^{\mathsf T}\frac{d}{dt}\phi_Q^t(z)
    = \nabla g(\phi_Q^t(z))^{\mathsf T}J\nabla Q(\phi_Q^t(z))
    = (L_Qg)(\phi_Q^t(z)),
\end{equation}
where the last equality uses $L_Qg=\{g,Q\}=\nabla g^{\mathsf T}J\nabla Q$.
The same identity with $g$ replaced by $L_Q^jg$ states that
\begin{equation*}
    \frac{d}{dt}(L_Q^jg)(\phi_Q^t(z))
    =\bigl(L_Q(L_Q^jg)\bigr)(\phi_Q^t(z))
    =(L_Q^{j+1}g)(\phi_Q^t(z)).
\end{equation*}
The base case is $L_Q^0g=g$. Induction on $j$ now yields
\begin{equation}
    \frac{d^j}{dt^j}g(\phi_Q^t(z))
    =(L_Q^jg)(\phi_Q^t(z)),\qquad j\geq0.
    \label{eq:flow-lie-derivatives}
\end{equation}

Since $\mathcal M_hz=\phi_F^h(\phi_G^h(z))$, Taylor's formula for
$t\mapsto f(\phi_F^t(\phi_G^h(z)))$ at $t=0$ through order $m$ is
\begin{equation}
    f(\mathcal M_hz)
    =\sum_{\beta=0}^m\frac{h^\beta}{\beta!}(L_F^\beta f)(\phi_G^h(z))
      +R_F(h,z),
    \label{eq:taylor-f-flow-expansion}
\end{equation}
where, by~\cref{eq:flow-lie-derivatives}, the integral remainder is
\begin{equation*}
    R_F(h,z)
    :=\frac{1}{m!}\int_0^h(h-t)^m
      (L_F^{m+1}f)(\phi_F^t(\phi_G^h(z)))\,dt.
\end{equation*}
By the assumed bound with $(\alpha,\beta)=(0,m+1)$, we have
\begin{equation}
    |R_F(h,z)|
    \leq\frac{M}{m!}\int_0^h(h-t)^m\,dt
    =\frac{Mh^{m+1}}{(m+1)!}.
    \label{eq:taylor-f-flow-remainder}
\end{equation}

Next, for each $\beta\in\{0,\ldots,m\}$, we apply Taylor's formula to
$t\mapsto(L_F^\beta f)(\phi_G^t(z))$ through order $m-\beta$.
By~\cref{eq:flow-lie-derivatives}, we obtain
\begin{equation}
    (L_F^\beta f)(\phi_G^h(z))
    =\sum_{\alpha=0}^{m-\beta}\frac{h^\alpha}{\alpha!}
      L_G^\alpha L_F^\beta f(z)+R_{G,\beta}(h,z),
    \label{eq:taylor-g-flow-expansion}
\end{equation}
where
\begin{equation*}
    R_{G,\beta}(h,z)
    :=\frac{1}{(m-\beta)!}\int_0^h(h-t)^{m-\beta}
      (L_G^{m+1-\beta}L_F^\beta f)(\phi_G^t(z))\,dt.
\end{equation*}
Since $(m+1-\beta) + \beta = m + 1$, the assumed bound yields
\begin{equation}
    \frac{h^\beta}{\beta!}|R_{G,\beta}(h,z)| \leq \frac{h^\beta}{\beta!} \frac{M}{(m-\beta)!}
      \int_0^h(h-t)^{m-\beta}\,dt
    = \frac{h^\beta}{\beta!} \frac{M h^{m-\beta+1}}{(m - \beta + 1)!}
    = \frac{Mh^{m+1}}{\beta!(m+1-\beta)!}.
    \label{eq:taylor-g-flow-remainder}
\end{equation}

We substitute~\cref{eq:taylor-g-flow-expansion} into~\cref{eq:taylor-f-flow-expansion}.
The triangle inequality then gives
\begin{align}
    \left|f(\mathcal M_hz)-\sum_{\alpha+\beta\leq m}
      \frac{h^{\alpha+\beta}}{\alpha!\beta!}
      L_G^\alpha L_F^\beta f(z)\right|
    & \leq |R_F(h,z)|
      +\sum_{\beta=0}^m\frac{h^\beta}{\beta!}|R_{G,\beta}(h,z)| \notag\\
    & \leq Mh^{m+1}\left(\frac{1}{(m+1)!}
      +\sum_{\beta=0}^m\frac{1}{\beta!(m+1-\beta)!}\right) \notag\\
    & =\frac{Mh^{m+1}}{(m+1)!}
      \sum_{\beta=0}^{m+1}\binom{m+1}{\beta}
      =\frac{2^{m+1}}{(m+1)!}Mh^{m+1},
\end{align}
where the second inequality uses~\cref{eq:taylor-f-flow-remainder,eq:taylor-g-flow-remainder}.
Since all bounds are uniform in $z$, we prove~\cref{eq:composed-taylor-remainder}.
\end{proof}

Finally,
we now combine the derivative bounds and Taylor formula to prove~\cref{lem:modified-energy-tool}.

\begin{proof}[Proof of~\cref{lem:modified-energy-tool}]
Fix an integer $N \geq 0$ and $0 < \eta\leq1/[16C_0(N+2)]$.
We first bound the difference between the truncated modified energy $\widehat{H}_\eta^{[N]}$ and $H_0$.
For $N=0$, this difference is zero because $\widehat H_\eta^{[0]}=H_0$,
so the first bound holds immediately.
For $N\geq1$ and $1\leq r\leq N$, \cref{lem:lie-derivative-bound} with $\alpha=\beta=0$ implies
\begin{equation}
    \|\eta^rH_r\|_\infty\leq \eta^r(4C_0)^{r+1}r! =: a_r.
    \label{eq:energy-correction-term-bound}
\end{equation}
For $1\leq r<N$, we have
\begin{equation}
    \frac{a_{r+1}}{a_r}
    =4\eta C_0(r+1)
    \leq\frac{r+1}{4(N+2)}\leq\frac14.
    \label{eq:energy-correction-ratio}
\end{equation}
Since $a_1=16C_0^2\eta$, by~\cref{eq:energy-correction-ratio}, we have $a_r\leq16C_0^2\eta(1/4)^{r-1}$.
By the triangle inequality and~\cref{eq:energy-correction-term-bound}, we obtain
\begin{equation}
    \|\widehat H_\eta^{[N]}-H_0\|_\infty \leq \sum_{r=1}^Na_r
    \leq 16C_0^2\eta\sum_{r=1}^N\left(\frac14\right)^{r-1}
    \leq \frac{64}{3}C_0^2\eta.
    \label{eq:truncated-energy-closeness}
\end{equation}
This proves the first bound in~\cref{eq:modified-energy-tool} for every $N\geq0$.

It remains to bound the change in $\widehat H_\eta^{[N]}$ over one iteration.
For each $r\in\{0,\ldots,N\}$ and $\alpha+\beta=N+2-r$, \cref{lem:lie-derivative-bound} implies
\begin{equation}
    \|L_G^\alpha L_F^\beta H_r\|_\infty
    \leq(4C_0)^{N+3}(N+2)!,
    \label{eq:energy-taylor-derivative-bound}
\end{equation}
since $r+\alpha+\beta=N+2\geq1$.
We apply~\cref{lem:real-taylor-composition} with $f=H_r$, $m=N+1-r$, and $M=(4C_0)^{N+3}(N+2)!$ to obtain
\begin{equation}
    H_r(\mathcal M_\eta z)
    =\sum_{\alpha+\beta\leq N+1-r}
    \frac{\eta^{\alpha+\beta}}{\alpha!\beta!}
    (L_G^\alpha L_F^\beta H_r)(z)+R_r(\eta,z),
    \label{eq:correction-taylor-expansion}
\end{equation}
where the remainder satisfies, for any $z$,
\begin{equation}
    \eta^r\|R_r(\eta, z)\|_\infty
    \leq\eta^{N+2}(4C_0)^{N+3}(N+2)!
    \frac{2^{N+2-r}}{(N+2-r)!}.
    \label{eq:weighted-correction-remainder}
\end{equation}

Note that the term with $\alpha=\beta=0$ in~\cref{eq:correction-taylor-expansion} is just $H_r(z)$.
We subtract this term from both sides of~\cref{eq:correction-taylor-expansion},
multiply by $\eta^r$ for each $r$,
and sum over $r=0,\ldots,N$. Thus
\begin{align*}
    \widehat H_\eta^{[N]}(\mathcal M_\eta z)-\widehat H_\eta^{[N]}(z)
    &=\sum_{r=0}^N\eta^r\bigl(H_r(\mathcal M_\eta z)-H_r(z)\bigr)\\
    &=\sum_{r=0}^N\sum_{1\leq\alpha+\beta\leq N+1-r}
    \frac{\eta^{r+\alpha+\beta}}{\alpha!\beta!}
    (L_G^\alpha L_F^\beta H_r)(z)
    +\sum_{r=0}^N\eta^rR_r(\eta,z).
\end{align*}
Equivalently, we have
\begin{equation}
    \widehat H_\eta^{[N]}(\mathcal M_\eta z)-\widehat H_\eta^{[N]}(z)
    =\sum_{j=1}^{N+1}\eta^j
    \sum_{r=0}^{j-1}\sum_{\alpha+\beta=j-r}
    \frac{(L_G^\alpha L_F^\beta H_r)(z)}{\alpha!\beta!}
    +\sum_{r=0}^N\eta^rR_r(\eta,z).
    \label{eq:truncated-energy-expansion}
\end{equation}

We next show that the first term in~\cref{eq:truncated-energy-expansion} vanishes.
For a formal variable $h$, the exponential operators in~\cref{eq:formal-update-operator} satisfy
\begin{equation*}
    e^{hL_G}e^{hL_F}-I
    =\sum_{\alpha+\beta\geq1}\frac{h^{\alpha+\beta}}{\alpha!\beta!}L_G^\alpha L_F^\beta,
\end{equation*}
where we subtract $I$ to remove the term $\alpha=\beta=0$.
This operator acts on $\widehat H_h^{[N]}=\sum_{r=0}^Nh^rH_r$ as follows:
\begin{equation*}
    \bigl( e^{hL_G}e^{hL_F} - I \bigr) \widehat{H}_h^{[N]}
    = \sum_{j \geq 1} h^j \sum_{\alpha + \beta + r = j, \alpha + \beta \geq 1, r \leq N} \frac{L_G^\alpha L_F^\beta H_r}{\alpha!\beta!}.
\end{equation*}

Consequently, for every $1\leq j\leq N+1$ and every $z$,
\begin{align*}
    \sum_{r=0}^{j-1}\sum_{\alpha+\beta=j-r}
    \frac{(L_G^\alpha L_F^\beta H_r)(z)}{\alpha!\beta!}
    &=[h^j]\Bigl(\bigl(e^{hL_G}e^{hL_F}-I\bigr)\widehat H_h^{[N]}\Bigr)(z)\\
    &=[h^j]\Bigl(\widehat H_h^{[N]}(\mathcal M_hz)-\widehat H_h^{[N]}(z)\Bigr)=0,
\end{align*}
where the second equality uses~\cref{eq:formal-update-operator}, and the last equality follows from~\cref{eq:finite-cancellation}:
the formal energy change is $O(h^{N+2})$, so all coefficients through degree $N+1$ vanish.
Therefore,~\cref{eq:truncated-energy-expansion} can be simplified to
\begin{equation}
    \widehat H_\eta^{[N]}(\mathcal M_\eta z)-\widehat H_\eta^{[N]}(z)
    =\sum_{r=0}^N\eta^rR_r(\eta,z).
    \label{eq:energy-defect-remainders}
\end{equation}

With the index $\ell=N+2-r$, the sum becomes
\begin{equation}
    \sum_{r=0}^N\frac{2^{N+2-r}}{(N+2-r)!}
    =\sum_{\ell=2}^{N+2}\frac{2^\ell}{\ell!}
    \leq\sum_{\ell=0}^{\infty}\frac{2^\ell}{\ell!}=e^2.
    \label{eq:remainder-factor-sum}
\end{equation}
By the triangle inequality and~\cref{eq:weighted-correction-remainder,eq:remainder-factor-sum},
the right-hand side of~\cref{eq:energy-defect-remainders} satisfies
\begin{align}
    \|\widehat H_\eta^{[N]}\circ\mathcal M_\eta-\widehat H_\eta^{[N]}\|_\infty
    &\leq\sum_{r=0}^N\eta^r\|R_r(\eta,\cdot)\|_\infty \notag\\
    &\leq \sum_{r=0}^N \frac{2^{N+2-r}}{(N+2-r)!} \eta^{N+2} (4C_0)^{N+3} (N+2)! \notag\\
    &\leq e^2\eta^{N+2}(4C_0)^{N+3}(N+2)!.
    \label{eq:raw-finite-defect}
\end{align}

Since $4\eta C_0\leq1/[4(N+2)]$ and $(N+2)! \leq (N+2)^{N+1}$, we can further simplify~\cref{eq:raw-finite-defect} using these bounds:
\begin{align}
    \|\widehat H_\eta^{[N]}\circ\mathcal M_\eta-\widehat H_\eta^{[N]}\|_\infty
    &\leq16e^2C_0^2\eta(4\eta C_0)^{N+1}(N+2)!\notag\\
    &\leq16e^2C_0^2\eta\,
    \frac{(N+2)!}{4^{N+1}(N+2)^{N+1}}\notag\\
    &\leq16e^2C_0^2\eta\,4^{-(N+1)}.
    \label{eq:truncated-energy-defect}
\end{align}
Together with~\cref{eq:truncated-energy-closeness}, we prove~\cref{lem:modified-energy-tool}.
\end{proof}

\subsection{Proof of the regret bound}
\label{app:regret-bound}

In this section, we first establish the comparator identity in~\cref{lem:regret-decomposition},
then use the modified-energy estimate to complete the proof of~\cref{thm:main}.

\begin{regretdecompositionrestatement}[Regret decomposition]
Fix a time horizon $T \geq 1$ and a step size $\eta > 0$.
For any comparators $(\tilde{x},\tilde{y}) \in \dom\psi_{\mathcal{X}}\times\dom\psi_{\mathcal{Y}}$,
define the Fenchel coupling:
\begin{equation*}
    \Gamma_{\tilde{x},\tilde{y}}(u,w)
    :=
    F(u) - \left( \ip{u}{\tilde{x}} - \psi_{\mathcal{X}}(\tilde{x}) \right)
    + G(w) - \left( \ip{w}{\tilde{y}} -  \psi_{\mathcal{Y}}(\tilde{y}) \right)
    \geq 0,
\end{equation*}
where $(u, w)$ are the dual variables in~\eqref{eq:altomd-dual}.
Then,
we have $\Gamma_{\tilde{x},\tilde{y}}(u_k, w_k) = D_{\psi_{\mathcal{X}}}(\tilde{x},x_k) + D_{\psi_{\mathcal{Y}}}(\tilde{y},y_k)$ for all $k \geq 0$.
Moreover, all the iterates generated by the AltOMD updates satisfy
\begin{equation*}
    \eta \bigl(
        \tilde{y}^\top A x_{k+1} - y_k^\top A \tilde{x}
    \bigr)
    =
    H_0(z_{k+1}) - H_0(z_k)
    + \Gamma_{\tilde{x},\tilde{y}}(z_k) - \Gamma_{\tilde{x},\tilde{y}}(z_{k+1}),
\end{equation*}
where $H_0(z) = F(u) + G(w)$ is defined in~\cref{eq:poisson-matrix}.
\end{regretdecompositionrestatement}

\paragraph{Convex duality identities.}
We prove the two identities needed to express the Fenchel coupling as a Bregman divergence.
\begin{proposition}[Fenchel--Legendre identities]
\label{prop:fenchel-legendre-identities}
Let $\psi:\R^d\to\R\cup\{+\infty\}$ be a proper convex function that is differentiable on $\ri(\dom\psi)$ relative to the affine hull of its domain.
Suppose $\psi^*$ is finite on a neighborhood of $u\in\R^d$ and differentiable at $u$,
and let $x=\nabla\psi^*(u)\in\ri(\dom\psi)$.
Then, for every $c\in\dom\psi$,
\begin{equation}
    \begin{aligned}
        \psi(x)+\psi^*(u) &= \ip{u}{x}, \\
        \ip{u-\nabla\psi(x)}{c-x} &= 0.
    \end{aligned}
\end{equation}
Here $\nabla\psi(x)$ is the gradient relative to the affine hull of $\dom\psi$.
\end{proposition}

\begin{proof}
For $c\in\dom\psi$, we differentiate the convexity inequality along
$x+t(c-x)$ at $t=0$ from the right. We obtain
\begin{equation*}
    \psi(c)\geq\psi(x)+\ip{\nabla\psi(x)}{c-x}.
\end{equation*}
Thus the supremum defining $\psi^*(\nabla\psi(x))$ is attained at $x$.
The same argument for the convex function $\psi^*$ at $u$, with
$\nabla\psi^*(u)=x$, implies
\begin{align*}
    \ip{\nabla\psi(x)}{x}-\psi(x)
    &=\psi^*(\nabla\psi(x))\\
    &\geq\psi^*(u)+\ip{x}{\nabla\psi(x)-u}.
\end{align*}
Equivalently, $\psi^*(u)\leq\ip{u}{x}-\psi(x)$.
The reverse inequality follows from the definition of $\psi^*(u)$,
so $\psi(x)+\psi^*(u)=\ip{u}{x}$.

Consequently, $x$ maximizes $\ip{u}{y}-\psi(y)$ over $y\in\dom\psi$.
Since $x\in\ri(\dom\psi)$, for every $c\in\dom\psi$ the points
$x+t(c-x)$ remain in the domain for all sufficiently small positive and negative $t$.
The derivative of $t\mapsto\ip{u}{x+t(c-x)}-\psi(x+t(c-x))$
therefore vanishes at $t=0$, so
$\ip{u-\nabla\psi(x)}{c-x}=0$.
\end{proof}

Fix an integer $T \geq 1$ and a constant stepsize $h > 0$,
and write $z_k = (u_k, w_k)$ for the dual iterates of~\cref{eq:altomd-dual} with $\eta = h$.
For comparators $\tilde{x}\in\dom\psi_{\mathcal{X}}$ and $\tilde{y}\in\dom\psi_{\mathcal{Y}}$,
use the Fenchel coupling $\Gamma_{\tilde{x},\tilde{y}}$ defined in~\cref{lem:regret-decomposition}.
We apply~\cref{prop:fenchel-legendre-identities} to $\psi_{\mathcal{X}}$ and
$\psi_{\mathcal{Y}}$. After we expand the Bregman divergences, we have
\begin{equation}
 \Gamma_{\tilde{x},\tilde{y}}(u,w)=D_{\psi_{\mathcal{X}}}(\tilde{x},\nabla F(u))
   +D_{\psi_{\mathcal{Y}}}(\tilde{y},\nabla G(w))\ge0.
\end{equation}

We next compute the change in the coupling over one iteration.
Because $F(u) + G(w) = H_0(z)$,
\begin{align}
    \Gamma_{\tilde{x},\tilde{y}}(z_{k+1})-\Gamma_{\tilde{x},\tilde{y}}(z_k)
    & = H_0(z_{k+1})-H_0(z_k) - \ip{u_{k+1}-u_k}{\tilde{x}}-\ip{w_{k+1}-w_k}{\tilde{y}} \notag \\
    & = H_0(z_{k+1})-H_0(z_k) - h\bigl(\tilde{y}^\top Ax_{k+1}-y_k^\top A\tilde{x}\bigr).
    \label{eq:comparator-identity-audited}
\end{align}
\Cref{eq:comparator-identity-audited} is equivalent to~\cref{eq:comparator-identity} with $\eta=h$.
By~\cref{ass:radius}, $\dom\psi_{\mathcal{X}}=\mathcal{X}$ and
$\dom\psi_{\mathcal{Y}}=\mathcal{Y}$, since any comparator outside an effective
domain would have infinite initial Bregman divergence.
Thus the identity applies to every comparator in $\mathcal{X}\times\mathcal{Y}$
in the regret argument below.
We sum over $k=0,\ldots,T-1$ and obtain
\begin{align*}
    h\sum_{k=0}^{T-1} \left( \tilde{y}^\top Ax_{k+1}-y_k^\top A\tilde{x} \right)
    &=H_0(z_T)-H_0(z_0)
    +\Gamma_{\tilde{x},\tilde{y}}(z_0)-\Gamma_{\tilde{x},\tilde{y}}(z_T)\\
    &\le H_0(z_T)-H_0(z_0)+\Lambda_0,
\end{align*}
where the inequality uses $\Gamma_{\tilde{x},\tilde{y}}(z_T)\ge0$ and
$\Gamma_{\tilde{x},\tilde{y}}(z_0)\le\Lambda_0$.

By~\cref{eq:regret}, the maximum over $(\tilde{x},\tilde{y})\in\mathcal{X}\times\mathcal{Y}$ satisfies
\begin{align}
    h\Reg_T
    &=\max_{(\tilde{x},\tilde{y})\in\mathcal{X}\times\mathcal{Y}}
    h\sum_{k=0}^{T-1}\bigl(\tilde{y}^\top Ax_{k+1}-y_k^\top A\tilde{x}\bigr)
    \notag\\
    &\leq\Lambda_0+H_0(z_T)-H_0(z_0).
    \label{eq:regret-energy-bound}
\end{align}

It remains to bound the change in $H_0 = F + G$ over the run.
Fix an integer $N\geq0$ and suppose $0<h\leq1/[16C_0(N+2)]$.
We first decompose the change in the Hamiltonian as follows.
\begin{equation}
    H_0(z_T)-H_0(z_0) =
    \bigl[H_0(z_T)-\widehat H_h^{[N]}(z_T)\bigr] + \bigl[\widehat H_h^{[N]}(z_T)-\widehat H_h^{[N]}(z_0)\bigr] + \bigl[\widehat H_h^{[N]}(z_0)-H_0(z_0)\bigr].
\end{equation}
By~\cref{lem:modified-energy-tool}, we have
\begin{align}
    H_0(z_T) - H_0(z_0)
    &\leq\frac{128}{3}C_0^2h+\sum_{k=0}^{T-1}\bigl|\widehat H_h^{[N]}(z_{k+1})-\widehat H_h^{[N]}(z_k)\bigr| \notag\\
    &\leq C_0^2h\left(\frac{128}{3}+16e^2T4^{-(N+1)}\right).
    \label{eq:change-in-hamiltonian}
\end{align}
From~\cref{eq:change-in-hamiltonian,eq:regret-energy-bound}, we obtain, after division by $h$,
\begin{equation}
    \Reg_T\leq\frac{\Lambda_0}{h}+C_0^2\left(\frac{128}{3}+16e^2T4^{-(N+1)}\right).
\end{equation}
Set $N=\lfloor\log_2T\rfloor$ and $h=\eta=1/[16C_0(N+2)]$ as in~\cref{thm:main}.
Since $T$ is an integer, $T\leq2^{N+1}-1\leq4^N$, where the second inequality follows from $(2^N-1)^2\geq0$.
Thus $T4^{-(N+1)}\leq1/4$, so
\begin{align*}
    \Reg_T
    &\leq16C_0\Lambda_0(N+2)+\left(\frac{128}{3}+4e^2\right)C_0^2\\
    &\leq16C_0\Lambda_0(\log_2T+2)+\left(\frac{128}{3}+4e^2\right)C_0^2.
\end{align*}
This proves~\cref{thm:main}.

\section{Verification of the entropy regularizer}
\label{app:entropy}

In this section, we verify the assumptions for negative entropy on the simplex and 
then prove the logarithmic regret bound in \cref{cor:entropy} for the alternating multiplicative weights update algorithm.

First, we record standard power-series estimates and Banach's theorem for symmetric multilinear forms.
\begin{proposition}
    \label{prop:power-series-polarization-input}
    For $r>0$, let $\mathcal A_r$ be the space of real power series
    $P(t)=\sum_{j\geq0}P_jt^j$ with finite weighted coefficient norm
    \begin{equation*}
        \norm{P}_r:=\sum\nolimits_{j\geq0}|P_j|r^j<\infty.
    \end{equation*}
    Then, the space $\mathcal A_r$ is complete, and
    \begin{equation}
        \norm{PQ}_r\leq\norm{P}_r\norm{Q}_r,
        \quad \forall \, P,Q \in \mathcal{A}_r.
        \label{eq:series-product-bound}
    \end{equation}
    For every $P \in \mathcal{A}_r$ and integer $s\geq0$, we have 
    \begin{equation}
        |P^{(s)}(0)|=s!|P_s|\leq s!r^{-s}\norm{P}_r.
        \label{eq:series-derivative-bound}
    \end{equation}
    Furthermore, for every integer $s\geq1$ and symmetric tensor $M$ of order $s$ on $\R^d$, we have 
    \begin{equation}
        \norm{M}_{\op}
        =\sup_{\norm{h}_2\leq1}|M[h,\ldots,h]|.
        \label{eq:real-polarization}
    \end{equation}
\end{proposition}

\begin{proof}[Proof of~\cref{prop:power-series-polarization-input}]
First, let $P^{[n]}(t)=\sum\nolimits_{j\geq0}a_{n,j}t^j$ be a Cauchy sequence
in $\mathcal A_r$. Since
$|a_{n,j}-a_{m,j}|r^j\leq\norm{P^{[n]}-P^{[m]}}_r$,
for each fixed $j$, the coefficient sequence converges to a real number
$a_j:=\lim_{n\to\infty}a_{n,j}$.
Define the candidate limit series $P_\infty(t):=\sum\nolimits_{j\geq0}a_jt^j$.
For every $\varepsilon>0$, choose $n_0$ such that
$\norm{P^{[n]}-P^{[m]}}_r\leq\varepsilon$ whenever $n,m\geq n_0$.
For each $n\geq n_0$ and integer $L\geq0$, coefficientwise convergence implies
\begin{equation*}
    \sum\nolimits_{j=0}^L|a_{n,j}-a_j|r^j
    =\lim_{m\to\infty}\sum\nolimits_{j=0}^L|a_{n,j}-a_{m,j}|r^j
    \leq\varepsilon.
\end{equation*}
We let $L\to\infty$ and conclude that $\norm{P^{[n]}-P_\infty}_r\leq\varepsilon$.
In particular, $\norm{P_\infty}_r\leq\norm{P^{[n_0]}}_r+\varepsilon<\infty$,
so $P_\infty\in\mathcal A_r$ and $P^{[n]}\to P_\infty$ in this space.
This proves completeness.

For $P,Q\in\mathcal A_r$, the coefficient of $t^j$ in $PQ$ is
$\sum\nolimits_{\ell=0}^jP_\ell Q_{j-\ell}$.
Hence, we have
\begin{align*}
    \norm{PQ}_r
    =\sum\nolimits_{j\geq0}
    \left|\sum\nolimits_{\ell=0}^jP_\ell Q_{j-\ell}\right|r^j
    \leq \sum\nolimits_{\ell\geq0}\sum\nolimits_{k\geq0}
    |P_\ell|r^\ell|Q_k|r^k
    =\norm{P}_r\norm{Q}_r,
\end{align*}
where the rearrangement is justified by Tonelli's theorem for nonnegative series.
This proves~\cref{eq:series-product-bound}.

Next, for $P\in\mathcal A_r$ and $|t|\leq r$, we have
$|P_jt^j|\leq |P_j|r^j$ and
$\sum_{j\geq0}|P_j|r^j=\norm{P}_r<\infty$.
Hence, the Weierstrass M-test implies that the series for $P$ converges absolutely and uniformly on $[-r,r]$.
For every $0<\rho<r$ and integer $s\geq1$, we have
\begin{equation*}
    \sum\nolimits_{j\geq s}\frac{j!}{(j-s)!}|P_j|\rho^{j-s}
    \leq\rho^{-s}\norm{P}_r
    \sup_{j\geq s}j^s\left(\frac{\rho}{r}\right)^j<\infty,
\end{equation*}
where the last inequality holds because geometric decay dominates the fixed power $j^s$.
Thus, every formally differentiated series converges uniformly on $[-\rho,\rho]$.
We integrate these series term by term and identify them successively as derivatives of $P$.
At zero, these derivatives satisfy $P^{(s)}(0)=s!P_s$.
Together with $|P_s|r^s\leq\norm{P}_r$, this proves~\cref{eq:series-derivative-bound} for $s\geq1$.
The case $s=0$ follows directly from $P(0)=P_0$.

Finally, \cref{eq:real-polarization} is by~\citet[Satz I]{banach1938homogene}
for symmetric multilinear forms on the real Hilbert space $(\R^d,\norm{\cdot}_2)$.
\end{proof}

\subsection{Shannon entropy regularizer}
\label{app:shannon}

\begin{proposition}
    \label{prop:shannon}
On $\Delta_d$, the function $\psi_d(x)=\sum_{i=1}^d x_i\log x_i$, with $0\log 0=0$, is Legendre relative to the affine hull and finite on the closed simplex.
As in~\cref{eq:shannon-mirror}, extend $\psi_d$ by $+\infty$ outside $\Delta_d$.
The conjugate of $\psi_d$ is
\begin{equation}
    F_d(u)=\log\sum_{i=1}^d e^{u_i},\qquad
    [\nabla F_d(u)]_i=\frac{e^{u_i}}{\sum_{j=1}^d e^{u_j}},
    \label{eq:shannon-conjugate}
\end{equation}
and, for every integer $s\geq1$,
\begin{equation}
    \sup_{u\in\R^d}\norm{D^sF_d(u)}_{\op}\leq\frac12\,4^s s!.
    \label{eq:shannon-derivative-bound}
\end{equation}
For each strictly positive $x\in\Delta_d$,
$\sup_{\tilde{x}\in\Delta_d}D_{\psi_d}(\tilde{x},x)
    \leq \log\frac{1}{\min_i x_i} < \infty$.
\end{proposition}

\begin{proof}
Since $(t\log t)''=1/t>0$ for $t>0$, $\psi_d$ is strictly convex and smooth
on the relative interior of $\Delta_d$, and it is finite and continuous on $\Delta_d$.
Convexity extends to the closed simplex by continuity.
Since $\Delta_d$ is closed and $\psi_d$ is continuous there, the epigraph
$\{(x,r):x\in\Delta_d,\ \psi_d(x)\leq r\}$ is closed and convex.
The extension by $+\infty$ is therefore proper, closed, and convex,
with effective domain $\Delta_d$.
The gradient of the entropy expression has coordinates $1+\log x_i$.
As $x$ approaches a relative boundary point, some $x_i\to0$ while another
$x_j$ tends to a positive value, so $\log x_i-\log x_j\to-\infty$.
After projection, the gradient relative to the affine hull preserves the coordinate differences, so the projected gradient also diverges.
Thus, the norm of the gradient relative to the affine hull tends to infinity,
so $\psi_d$ has the required Legendre property.

Next, fix $u\in\R^d$ and set $p_i=e^{u_i}/\sum\nolimits_{j=1}^d e^{u_j}$.
For every $x\in\Delta_d$, the identity $\sum\nolimits_{i=1}^d x_i=1$ implies
\begin{align}
    \ip{u}{x}-\psi_d(x)
    &=\sum\nolimits_{i=1}^d x_i\left(\log p_i+\log\sum\nolimits_{j=1}^d e^{u_j}-\log x_i\right) \notag\\
    &=\log\sum\nolimits_{j=1}^d e^{u_j}-\sum\nolimits_{i=1}^d x_i\log\frac{x_i}{p_i},
    \label{eq:shannon-conjugate-objective}
\end{align}
where terms with $x_i=0$ are interpreted by continuity.
We apply $-\log a\geq1-a$ with $a=p_i/x_i$ on the positive coordinates of $x$ and obtain
\begin{equation*}
    \sum\nolimits_{i=1}^d x_i\log\frac{x_i}{p_i}
    \geq\sum\nolimits_{\substack{1\leq i\leq d\\x_i>0}}x_i\left(1-\frac{p_i}{x_i}\right)
    =1-\sum\nolimits_{\substack{1\leq i\leq d\\x_i>0}}p_i\geq0.
\end{equation*}
The sum $\sum\nolimits_{i=1}^d x_i\log\frac{x_i}{p_i}$ vanishes at $x=p$.
We take the supremum over $x \in \Delta_d$ in~\cref{eq:shannon-conjugate-objective}
to obtain the conjugate in~\cref{eq:shannon-conjugate}.
We differentiate $F_d(u) = \log\sum\nolimits_{j=1}^d e^{u_j}$ to obtain the gradient in~\cref{eq:shannon-conjugate} and verify that $F_d$ is smooth on $\R^d$.

To prove~\cref{eq:shannon-derivative-bound}, fix $h\in\R^d$ with $\norm{h}_2\leq1$.
In terms of the probabilities $p_i$ defined above, we can write
\begin{align}
    F_d(u+th)-F_d(u)
    &=\log\left(\frac{\sum\nolimits_{i=1}^d e^{u_i}e^{th_i}}{\sum\nolimits_{j=1}^d e^{u_j}}\right) \notag\\
    &=\log\left(\sum\nolimits_{i=1}^d p_i e^{th_i}\right)
    =\log(1+P(t)),
    \label{eq:shannon-directional-increment}
\end{align}
where $P(t):=\sum\nolimits_{i=1}^d p_i(e^{th_i}-1)$.
We expand the exponentials as follows:
\begin{equation}
    P(t) = \sum\nolimits_{i=1}^d p_i \left( \sum\nolimits_{j\geq1}\frac{(th_i)^j}{j!} \right)
    = \sum\nolimits_{j\geq1}\left(\frac{1}{j!}\sum\nolimits_{i=1}^d p_i h_i^j\right)t^j.
\end{equation}
Here, the exchange of sums is valid because the sum over $i$ is finite and each
exponential series converges absolutely for every real $t$.
Let $\norm{P}_q:=\sum_{j\geq0}|P_j|q^j$ for $q>0$.
Since $|h_i|\leq\norm{h}_2\leq1$ and $\sum\nolimits_{i=1}^d p_i=1$,
\begin{equation}
    \norm{P}_{1/4} = \sum\nolimits_{j\geq1}\left|\sum\nolimits_{i=1}^d p_i h_i^j\right|\frac{1}{4^j j!}
    \leq \sum\nolimits_{i=1}^d p_i\sum\nolimits_{j\geq1}\frac{|h_i|^j}{4^j j!}
    \leq \sum\nolimits_{j\geq1} \frac{1}{4^j j!}
    \leq \sum\nolimits_{j\geq1} \frac{1}{4^j} = \frac{1}{3}.
    \label{eq:shannon-increment-norm}
\end{equation}
By~\cref{eq:series-product-bound}, the logarithm series
\begin{equation}
    \log(1+P):=\sum\nolimits_{k\geq1}\frac{(-1)^{k+1}}{k}P^k
    \label{eq:shannon-logarithm-series}
\end{equation}
converges in $\mathcal A_{1/4}$, with
\begin{equation}
    \norm{\log(1+P)}_{1/4}
    \leq\sum\nolimits_{k\geq1}\frac{\norm{P^k}_{1/4}}{k}
    \leq\sum\nolimits_{k\geq1}\frac{\norm{P}_{1/4}^k}{k}
    \leq\sum\nolimits_{k\geq1}\frac{3^{-k}}{k}=\log\frac32<\frac12.
    \label{eq:shannon-logarithm-norm}
\end{equation}
For real $|t|\leq1/4$, we have $|P(t)|\leq\norm{P}_{1/4}<1$,
so the logarithm series in~\cref{eq:shannon-logarithm-series}
represents the real logarithm in~\cref{eq:shannon-directional-increment}.
For every integer $s\geq1$, \cref{eq:series-derivative-bound} implies
\begin{align}
    |D^sF_d(u)[h,\ldots,h]|
    &=\Bigg| \ \left.\frac{d^s}{dt^s}
    \bigl(F_d(u+th)-F_d(u)\bigr)\right|_{t=0} \ \Bigg| \notag\\
    &=s!\left|[t^s]\log(1+P(t))\right|
    \leq4^s s!\norm{\log(1+P)}_{1/4}
    \leq\frac12\,4^s s!.
    \label{eq:shannon-directional-bound}
\end{align}
Since $D^sF_d(u)$ is a symmetric tensor of order $s$,~\cref{eq:real-polarization}
now implies~\cref{eq:shannon-derivative-bound}:
\begin{equation*}
    \norm{D^sF_d(u)}_{\op}
    =\sup_{\norm{h}_2\leq1}|D^sF_d(u)[h,\ldots,h]|
    \leq\frac12\,4^s s!.
\end{equation*}

Finally, fix a strictly positive $x\in\Delta_d$.
The relative gradient is the orthogonal projection of $(1+\log x_i)_{i=1}^d$
onto $\{v:\sum_i v_i=0\}$.
Since $\sum_i(\tilde{x}_i-x_i)=0$ for every $\tilde{x}\in\Delta_d$,
this projection does not change the inner product with $\tilde{x}-x$,
so we may use the vector $(1+\log x_i)_{i=1}^d$ in the Bregman formula.
For every $\tilde{x}\in\Delta_d$, the Bregman divergence has the expansion
\begin{align}
    D_{\psi_d}(\tilde{x},x)
    &=\sum\nolimits_{i=1}^d\tilde{x}_i\log\tilde{x}_i-\sum\nolimits_{i=1}^d x_i\log x_i
    -\sum\nolimits_{i=1}^d(1+\log x_i)(\tilde{x}_i-x_i) \notag\\
    &=\sum\nolimits_{i=1}^d\tilde{x}_i\log\frac{\tilde{x}_i}{x_i},
    \label{eq:shannon-bregman-divergence}
\end{align}
where the second equality uses $\sum\nolimits_{i=1}^d\tilde{x}_i=\sum\nolimits_{i=1}^d x_i=1$.
Since $0\leq\tilde{x}_i\leq1$, we have $\sum\nolimits_{i=1}^d\tilde{x}_i\log\tilde{x}_i\leq0$, and hence
\begin{equation*}
    D_{\psi_d}(\tilde{x},x)
    \leq\sum\nolimits_{i=1}^d\tilde{x}_i\log\frac{1}{x_i}
    \leq\left(\sum\nolimits_{i=1}^d\tilde{x}_i\right)\log\frac{1}{\min_i x_i}
    =\log\frac{1}{\min_i x_i}.
\end{equation*}
This completes~\cref{prop:shannon}.
\end{proof}

\begin{entropyrestatement}[Alternating multiplicative weights update]
    Let $A\in\R^{m\times n}$ and let $x_0\in\Delta_n$ and $y_0\in\Delta_m$
    be strictly positive initial strategies.
    Set
    $C = 384\max\{\norm{A}, 1\}$
    and $\Lambda_0 = \log({1}/{\min_{1\leq i\leq n}x_{0,i}}) + \log({1}/{\min_{1\leq j\leq m}y_{0,j}})$.
    For every integer $T\geq1$,
    choose the constant step size $\eta = \frac{1}{16C(\lfloor\log_2T\rfloor+2)}$,
    and generate a sequence of iterates for $k=0,\ldots,T-1$ by the updates in~\cref{eq:entropy}.
    Then, the total regret satisfies
    $\Reg_T \leq 16 C \Lambda_0(\log_2T+2)+\left(\frac{128}{3}+4e^2\right)C^2$.
    In particular, the regret bound for uniform initial strategies is
    \begin{equation*}
        \Reg_T \leq 16C\log(nm)(\log_2T+2)+\left(\frac{128}{3}+4e^2\right)C^2.
    \end{equation*}
\end{entropyrestatement}

\begin{proof}[Proof of Corollary~\ref{cor:entropy}]
By~\cref{prop:shannon}, the two entropy generators satisfy the relative Legendre assumptions.
We apply~\cref{eq:shannon-derivative-bound} to $F=F_n$ and $G=F_m$.
For every integer $s\geq1$, we obtain
\begin{equation*}
    \frac{1}{s!}\left(
    \sup_u\norm{D^sF(u)}_{\op}+\sup_w\norm{D^sG(w)}_{\op}\right)
    \leq2\cdot\frac12\,4^s=4^s.
\end{equation*}
Thus~\cref{ass:regularity} holds with $K=4$.
Here $\norm{A}$ denotes the spectral norm.
Since $J_A^\top J_A=\operatorname{diag}(A^\top A,AA^\top)$,
we have $\norm{J_A}_{\op}=\norm{A}$.
The constant in~\cref{lem:modified-energy-tool} therefore becomes
\begin{equation*}
    C_0=24\cdot4^2\max\{\norm{A},1\}
    =384\max\{\norm{A},1\}=C,
\end{equation*}
as stated in the corollary.
For strictly positive initial strategies,~\cref{prop:shannon} yields
\begin{equation*}
    \sup_{\tilde{x}\in\Delta_n}D_{\psi_n}(\tilde{x},x_0)
    +\sup_{\tilde{y}\in\Delta_m}D_{\psi_m}(\tilde{y},y_0)
    \leq\log\frac{1}{\min_{1\leq i\leq n}x_{0,i}}
    +\log\frac{1}{\min_{1\leq j\leq m}y_{0,j}}
    =\Lambda_0<\infty,
\end{equation*}
which verifies~\cref{ass:radius}, with initial comparator radius at most
the stated $\Lambda_0$.

It remains to identify the updates with~\cref{eq:entropy}.
Since $x_{k,i}=e^{u_{k,i}}/\sum\nolimits_{\ell=1}^n e^{u_{k,\ell}}$,
the first update in~\cref{eq:altomd-dual} satisfies
\begin{equation*}
    x_{k+1,i} = \frac{e^{u_{k,i}-\eta(A^\top y_k)_i}}
    {\sum\nolimits_{\ell=1}^n e^{u_{k,\ell}-\eta(A^\top y_k)_\ell}} = \frac{x_{k,i}e^{-\eta(A^\top y_k)_i}}
    {\sum\nolimits_{\ell=1}^n x_{k,\ell}e^{-\eta(A^\top y_k)_\ell}},
\end{equation*}
where we divided the numerator and denominator by $\sum\nolimits_{\ell=1}^n e^{u_{k,\ell}}$.
The same calculation for $w_{k+1}=w_k+\eta Ax_{k+1}$
establishes the second update in~\cref{eq:entropy}.
Set $N=\lfloor\log_2T\rfloor$. The stated step size is
$\eta=1/[16C(N+2)]=1/[16C_0(N+2)]$, as required in~\cref{thm:main}.
With the initial comparator radius bounded by $\Lambda_0$, \cref{thm:main} implies
\begin{equation*}
    \Reg_T\leq16C\Lambda_0(N+2)+\left(\frac{128}{3}+4e^2\right)C^2
    \leq16C\Lambda_0(\log_2T+2)+\left(\frac{128}{3}+4e^2\right)C^2.
\end{equation*}
For uniform initialization, we have $x_{0,i}=1/n$ and $y_{0,j}=1/m$,
so $\Lambda_0=\log n+\log m=\log(nm)$. This proves~\cref{eq:entropy-regret}.
\end{proof}

\end{document}